%% file: camera_ready.tex
\documentclass{article}

\PassOptionsToPackage{numbers}{natbib}
\usepackage[final]{neurips_2023}

\usepackage[utf8]{inputenc} 
\usepackage[T1]{fontenc}    
\usepackage{hyperref}       
\usepackage{url}            
\usepackage{booktabs}       
\usepackage{amsfonts}       
\usepackage{nicefrac}       
\usepackage{microtype}      
\usepackage{xcolor}         

\usepackage{graphicx}
\usepackage{subfig}
\usepackage{subfloat}
\usepackage{amsmath,amssymb}
\usepackage{subfig}
\usepackage{mathtools}
\usepackage{algorithm}
\usepackage{algorithmic}
\usepackage{amsthm}
\usepackage{booktabs}  
\usepackage{arydshln}
\usepackage{subfig}
\usepackage{multirow}
\usepackage{enumitem}
\usepackage{wrapfig}
\usepackage{fancyhdr}
\usepackage{color}
\usepackage{colortbl}

\usepackage{bm}

\usepackage{listings}

\title{Diffusion Models and Semi-Supervised Learners Benefit Mutually with Few Labels}

\author{%
  Zebin You$^{1, 2}$\thanks{Equal contribution.}, Yong Zhong$^{1, 2}$\footnotemark[1], Fan Bao$^3$, Jiacheng Sun$^{4}$, Chongxuan Li$^{1,2 }$\thanks{Correspondence to Chongxuan Li.}, Jun Zhu$^{3}$\\
  $^1$ Gaoling School of Artificial Intelligence, Renmin University of China, Beijing, China\\
  $^2$ Beijing Key Laboratory of Big Data Management and Analysis Methods, Beijing, China\\
  $^3$ Dept. of Comp. Sci. \& Tech., BNRist Center, THU-Bosch ML Center, Tsinghua University \\
  $^4$ Huawei Noah’s Ark Lab \\
  \texttt{zebin@ruc.edu.cn;} \texttt{yongzhong@ruc.edu.cn;} \texttt{bf19@mails.tsinghua.edu.cn;} \\
  \texttt{sunjiacheng1@huawei.com;} \texttt{chongxuanli@ruc.edu.cn;} \texttt{dcszj@tsinghua.edu.cn} 
}

\begin{document}
\maketitle

\begin{abstract}
In an effort to further advance semi-supervised generative and classification tasks, we propose a simple yet effective training strategy called \emph{dual pseudo training} (DPT), built upon strong semi-supervised learners and diffusion models. DPT operates in three stages: training a classifier on partially labeled data to predict pseudo-labels; training a conditional generative model using these pseudo-labels to generate pseudo images; and retraining the classifier with a mix of real and pseudo images.  Empirically, DPT consistently achieves SOTA performance of semi-supervised generation and classification across various settings. In particular, with one or two labels per class, DPT achieves a Fréchet Inception Distance (FID) score of 3.08 or 2.52 on ImageNet $256\times256$. Besides, DPT outperforms competitive semi-supervised baselines substantially on  ImageNet classification tasks, \emph{achieving top-1 accuracies of 59.0 (+2.8), 69.5 (+3.0), and 74.4 (+2.0)} with one, two, or five labels per class, respectively. Notably, our results demonstrate that diffusion can generate realistic images with only a few labels (e.g., $<0.1\%$) and generative augmentation remains viable for semi-supervised classification. Our code is available at \emph{\href{https://github.com/ML-GSAI/DPT}{https://github.com/ML-GSAI/DPT}}.
\end{abstract}

\section{Introduction}
\label{sec:intro}

Diffusion probabilistic models~\citep{sohl2015deep,ho2020denoising,song2020score,dhariwal2021diffusion,bao2022all,Karras2022edm,peebles2022scalable} have achieved excellent performance in image generation. However, empirical evidence has shown that labeled data is indispensable for training such models~\citep{bao2022conditional,dhariwal2021diffusion}. Indeed, lacking labeled data leads to much lower performance of the generative model. For instance, the representative work (i.e., ADM)~\citep{dhariwal2021diffusion} achieves an FID of 10.94 on fully labeled ImageNet $256\times256$, while an FID of 26.21 without labels. 

To improve the performance of diffusion models without utilizing labeled data, prior work~\citep{bao2022conditional, hu2023self} initially conducts clustering and subsequently trains diffusion models conditioned on the cluster indices. Although these methods can, in some instances, exhibit superior performance over supervised models on low-resolution data, such phenomena have not yet been observed on high-resolution data (e.g., on ImageNet $256\times256$, an FID of 5.19, compared to an FID of 3.31 achieved by supervised models, see Appendix~\ref{app:model_arc}). Besides, cluster indices may not always align with ground truth labels, making it hard to control semantics in samples. 
\begin{figure*}[!htb]
\begin{center}
\includegraphics[width=1.0\linewidth]{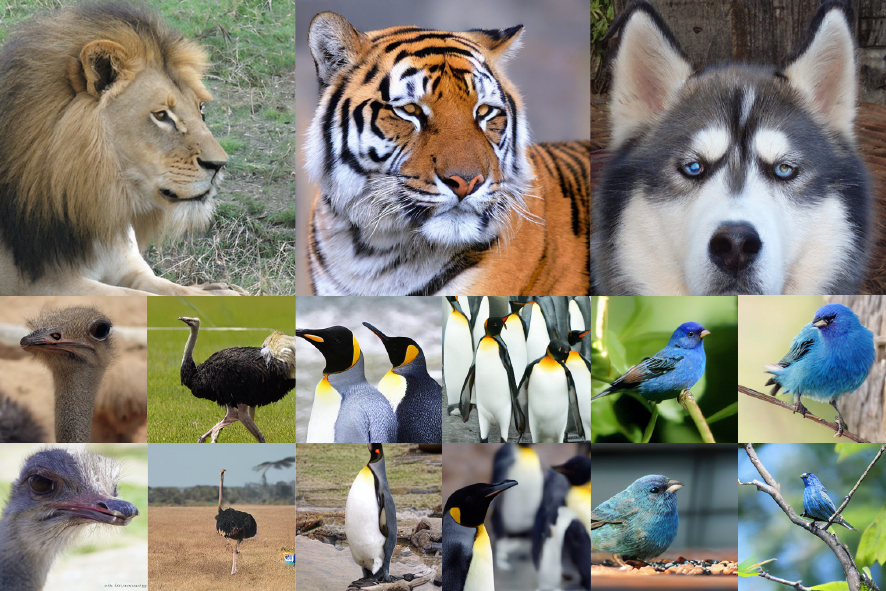}
\end{center}
\caption{Selected samples from DPT. Top row: $512\times512$ samples from DPT trained with \textbf{five ($<$ 0.4\%)} labels per class. Bottom rows: $256\times256$ samples from DPT trained with \textbf{one ($<$ 0.1\%)} label per class ~(\emph{Left:} ``Ostrich''; \emph{Mid:} ``King penguin''; \emph{Right:} ``Indigo bunting'').}
\label{fig:gen_summary}
\vspace{-.3cm}
\end{figure*}
Compared to unsupervised methods, semi-supervised generative models~\citep{kingma2014semi,li2017triple,luvcic2019high} often perform much better and provide the same way to control the semantics of samples as the supervised ones by using a small number of labels.  
However, to our knowledge, although it is attractive, little work in the literature has investigated semi-supervised diffusion models. This leads us to a key question: can diffusion models generate high-fidelity images with controllable semantics given only a few (e.g., $<0.1 \%$) labels?

On the other hand, while it is natural to use images sampled from generative models for semi-supervised classification~\citep{kingma2014semi,li2017triple}, discriminative methods~\citep{sohn2020fixmatch, zhang2021flexmatch, wang2022freematch} dominant the area recently. In particular, self-supervised based learners~\citep{cai2022semi, assran2022masked, chen2020big} have demonstrated state-of-the-art performance on ImageNet. However, generative models have rarely been considered for semi-supervised classification recently. Therefore another key question arises: can generative augmentation be a useful approach for such strong semi-supervised classifiers, with the aid of advanced diffusion models?

To answer the above two key and pressing questions, we propose a simple but effective training strategy called \emph{dual pseudo training} (DPT), built upon strong diffusion models and semi-supervised classifiers.
DPT is three-staged (see Fig.~\ref{fig:flow_chart}). First, a classifier is trained on partially labeled data and used to predict pseudo-labels for all data. Second, a conditional generative model is trained on all data with pseudo-labels and used to generate pseudo images given labels. Finally, the classifier is trained on real data augmented by pseudo images with labels. Intuitively, in DPT, the two opposite conditional models (i.e. diffusion model and classifier) provide complementary learning signals to each other and benefit mutually (see a detailed discussion in Appendix~\ref{app:thought_experiment}).

\begin{figure}[t!]
\begin{center}
\subfloat[DPT vs. supervised diffusion models. 
]{\includegraphics[width=.485\columnwidth]{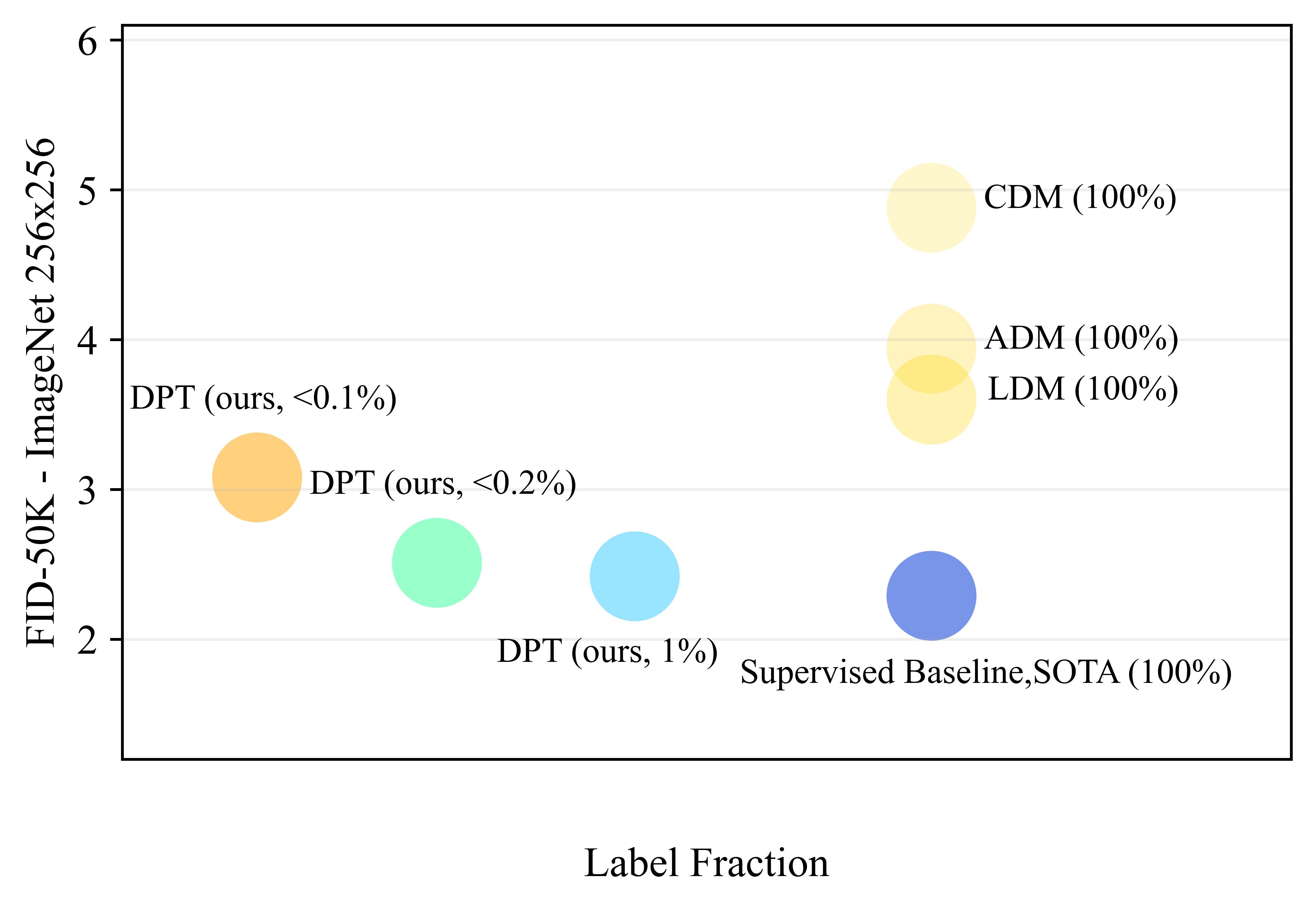}}
\subfloat[DPT vs. semi-supervised classifiers.]{\includegraphics[width=.485\columnwidth]{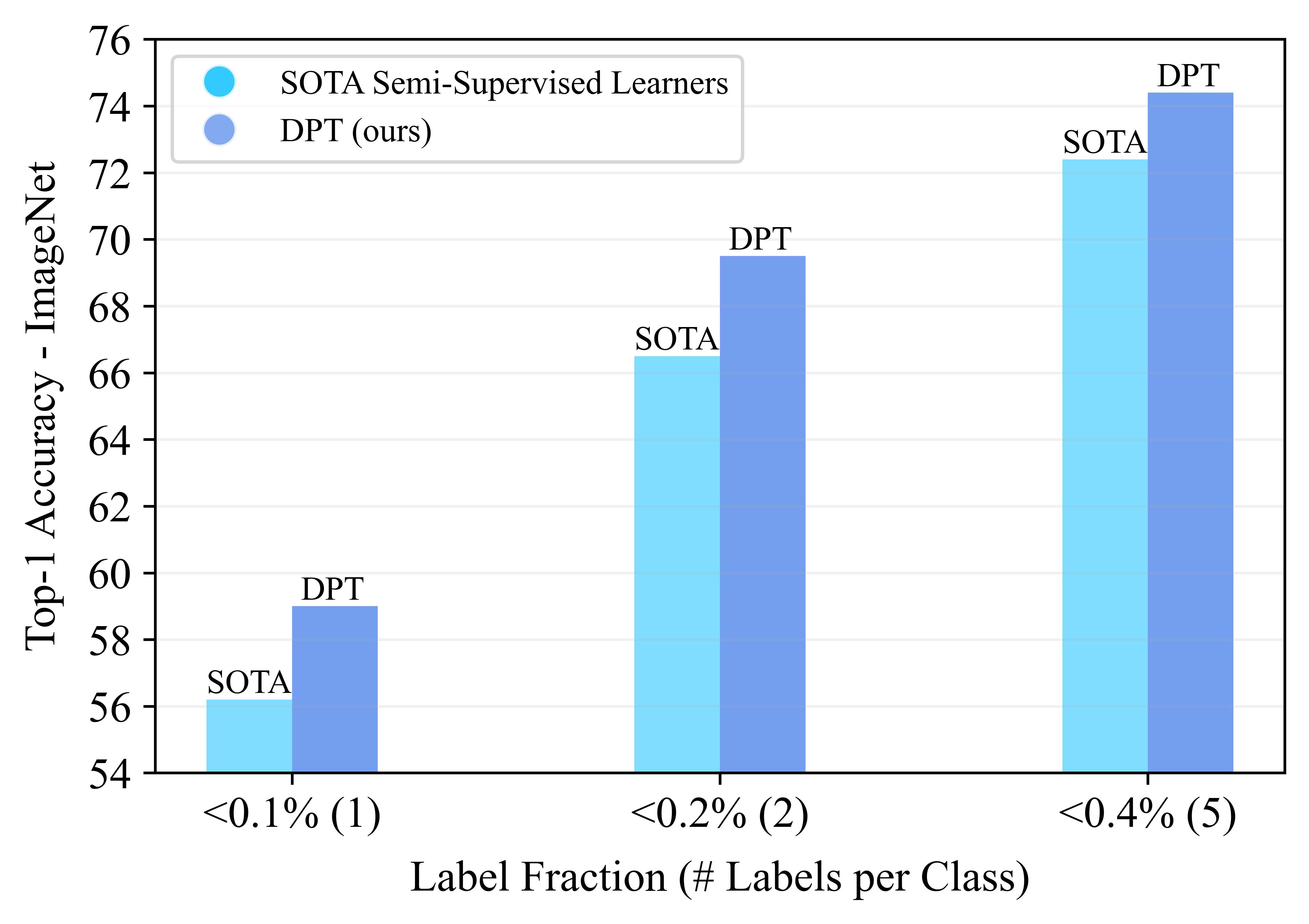}}
\end{center}
\vspace{-.2cm}
\caption{\textbf{Generation and classification results of DPT on ImageNet with few labels.} (a) DPT with $<$ 0.1\% labels outperforms strong supervised diffusion models~\citep{dhariwal2021diffusion,ho2022cascaded,rombach2022high}. (b) DPT substantially improves SOTA semi-supervised learners~\citep{assran2022masked}.}
\label{fig:accuracy}
\vspace{-.3cm}
\end{figure}

We evaluate the effectiveness of DPT through diverse experiments on multi-scale and multi-resolution benchmarks, including CIFAR-10~\citep{krizhevsky2009learning} and ImageNet~\citep{imagenet} at resolutions of $128\times128$, $256\times256$, and $512\times512$. Quantitatively, DPT obtains SOTA semi-supervised generation results on two common metrics, including FID~\citep{heusel2017gans} and IS~\citep{salimans2016improved}, in all settings. In particular, in the highly appealing task, i.e. ImageNet $256\times256$ generation, DPT with \emph{one} (i.e., $<$ 0.1\%) labels per class achieves an FID of 3.08, outperforming strong supervised diffusion models including IDDPM~\citep{nichol2021improved}, CDM~\citep{ho2022cascaded}, ADM~\citep{dhariwal2021diffusion} and LDM~\citep{rombach2022high} (see Fig.~\ref{fig:accuracy} (a)). It is worth noting that the comparison with previous models here is meant to illustrate that DPT maintains good performance even with minimal labels, rather than directly comparing it to these previous models (direct comparison is unfair as different diffusion models were used). Furthermore, DPT with \emph{two} (i.e., $<$ 0.2\%) labels per class is comparable to supervised baseline U-ViT~\citep{bao2022all} (FID 2.52 vs. 2.29). Moreover, on ImageNet $128\times128$ generation, DPT with \emph{one} (i.e., $<$ 0.1 \%) labels per class outperforms SOTA semi-supervised generative models S$^3$GAN~\citep{luvcic2019high} with 20$\%$ labels (FID 4.59 vs. 7.7).
Qualitatively, DPT can generate realistic, diverse, and semantically correct images with very few labels, as shown in Fig~\ref{fig:gen_summary}. We also explore why classifiers can benefit generative models through class-level visualization and analysis in Appendix~\ref{app:class_wise_analysis_classification_help_generation}. 

As for semi-supervised classification, DPT achieves state-of-the-art (SOTA) performance in various settings, including ImageNet with one, two, five labels per class and 1$\%$ labels. On the smaller dataset, namely CIFAR-10, DPT with four labels per class achieves the second-best error rate of 4.68{\scriptsize $\pm$0.17}\%.  Besides, on ImageNet classification benchmarks with one, two, five labels per class and 1$\%$ labels, DPT outperforms competitive semi-supervised baselines~\citep{assran2022masked, cai2022semi}, achieving state-of-the-art top-1 accuracy of 59.0 (+2.8), 69.5 (+3.0), 74.4 (+2.0) and 80.2 (+0.8) respectively (see Fig.~\ref{fig:accuracy} (b)). Similarly to generation tasks, we also investigate
why generative models can benefit classifiers via class-level visualization and analysis in Appendix ~\ref{app:analysis_generation_help_classification}. 

In summary, our novelty and key contributions are as follows:
\begin{itemize}[leftmargin=*]
\setlength\itemindent{.5em}
\item We present Dual Pseudo Training (DPT), a straightforward yet effective strategy designed to advance the frontiers of semi-supervised diffusion models and classifiers.
\item We achieve SOTA semi-supervised generation performance on CIFAR-10 and ImageNet datasets across various settings. Moreover, we demonstrate that diffusion models with a few labels (e.g., $<0.1\%$) can generate realistic, diverse, and semantically accurate images, as depicted in Fig~\ref{fig:gen_summary}.
\item We achieve SOTA semi-supervised classification performance on ImageNet datasets across various settings and the second-best results on CIFAR-10. Besides, we demonstrate that aided by diffusion models, generative augmentation remains a viable approach for semi-supervised classification.
\item We explore why diffusion models and semi-supervised learners benefit mutually with few labels via class-level visualization and analysis, as showcased in Appendix~\ref{app:class_wise_analysis_classification_help_generation} and Appendix ~\ref{app:analysis_generation_help_classification}. 
\end{itemize}

\section{Settings and Preliminaries}
\label{sec:preliminary}

We present settings and preliminaries on two representative self-supervised based learners for semi-supervised learning~\citep{assran2022masked}~\citep{cai2022semi} in Sec.~\ref{sec:msn_semi_vit} and conditional diffusion probabilistic  models~\citep{ho2020denoising,bao2022all,ho2022classifier} in Sec.~\ref{sec:ddpm}, respectively. We consider image generation and classification in semi-supervised learning, where the training set consists of $N$ labeled images $\mathcal{S} = \{(\mathbf{x}_i^l,y_i^l)\}_{i=1}^N$ and $M$ unlabeled images $\mathcal{D} = \{\mathbf{x}_i^u\}_{i=1}^M$. We assume $N \ll M$. For convenience, we denote the set of all real images as $\mathcal{X} = \{\mathbf{x}_i^u\}_{i=1}^{M} \cup \{\mathbf{x}_i^l\}_{i=1}^{N}$, and the set of all possible classes as $\mathcal{Y}$.

\subsection{Semi-Supervised Classifier}
\label{sec:msn_semi_vit} 

\textbf{Masked Siamese Networks (MSN)}~\citep{assran2022masked} employ a ViT-based~\citep{DBLP:conf/iclr/DosovitskiyB0WZ21} anchor encoder $f_{\bm{\theta}}(\cdot)$ and a target encoder $f_{\bm{\bar{\theta}}}(\cdot)$, 
where $\bm{\bar{\theta}}$ is the exponential moving average (EMA)~\citep{grill2020bootstrap} of parameters $\bm\theta$.
For a real image $\mathbf{x}_i \in \mathcal{X}, 1\le i \le M+N$, MSN obtains $H+1$ random augmented images, denoted as 
$\mathbf{x}_{i,h}, 1\le h \le H+1$. 
MSN then applies either a random mask or a focal mask to the first $H$ augmented images and obtain $\textrm{mask}(\mathbf{x}_{i,h}), 1\le h \le H$. MSN optimizes $\bm\theta$ and a learnable matrix of prototypes $\mathbf{q}$ by the following objective function:
\begin{align}
    \frac{1}{H(M+N)}\sum_{i=1}^{M+N}\sum_{h=1}^H \textrm{CE}(\mathbf{p}_{i,h},\mathbf{p}_{i, H+1}) - \lambda \textrm{H}(\bar{\mathbf{p}}),
    \label{eq:msn}
\end{align}
where $\textrm{CE}$ and $\textrm{H}$ are cross entropy and entropy respectively, $\mathbf{p}_{i,h} = \textrm{softmax}( (f_{\bm\theta}(\textrm{mask}(\mathbf{x}_{i,h})) \cdot \mathbf{q} / \tau)$ , $\mathbf{\bar{p}}$ is the mean of $\mathbf{p}_{i,h}$, $\mathbf{p}_{i, H+1} = \textrm{softmax}( f_{\bm{\bar{{\theta}}}}(\mathbf{x}_{i, H+1}) \cdot \mathbf{q} / \tau')$, $\tau$,$\tau'$ and $\lambda$ are hyper-parameters, and $\cdot$ denotes cosine similarity. MSN is an efficient semi-supervised approach by extracting features for all labeled images in $\mathcal{S}$ and training a linear classifier on top of the features using L$_2$-regularized logistic regression. When a self-supervised pre-trained model is available, MSN demonstrates high efficiency in training a semi-supervised classifier on a single CPU core.

\begin{figure*}[t!]
\centering    \includegraphics[width=\linewidth]{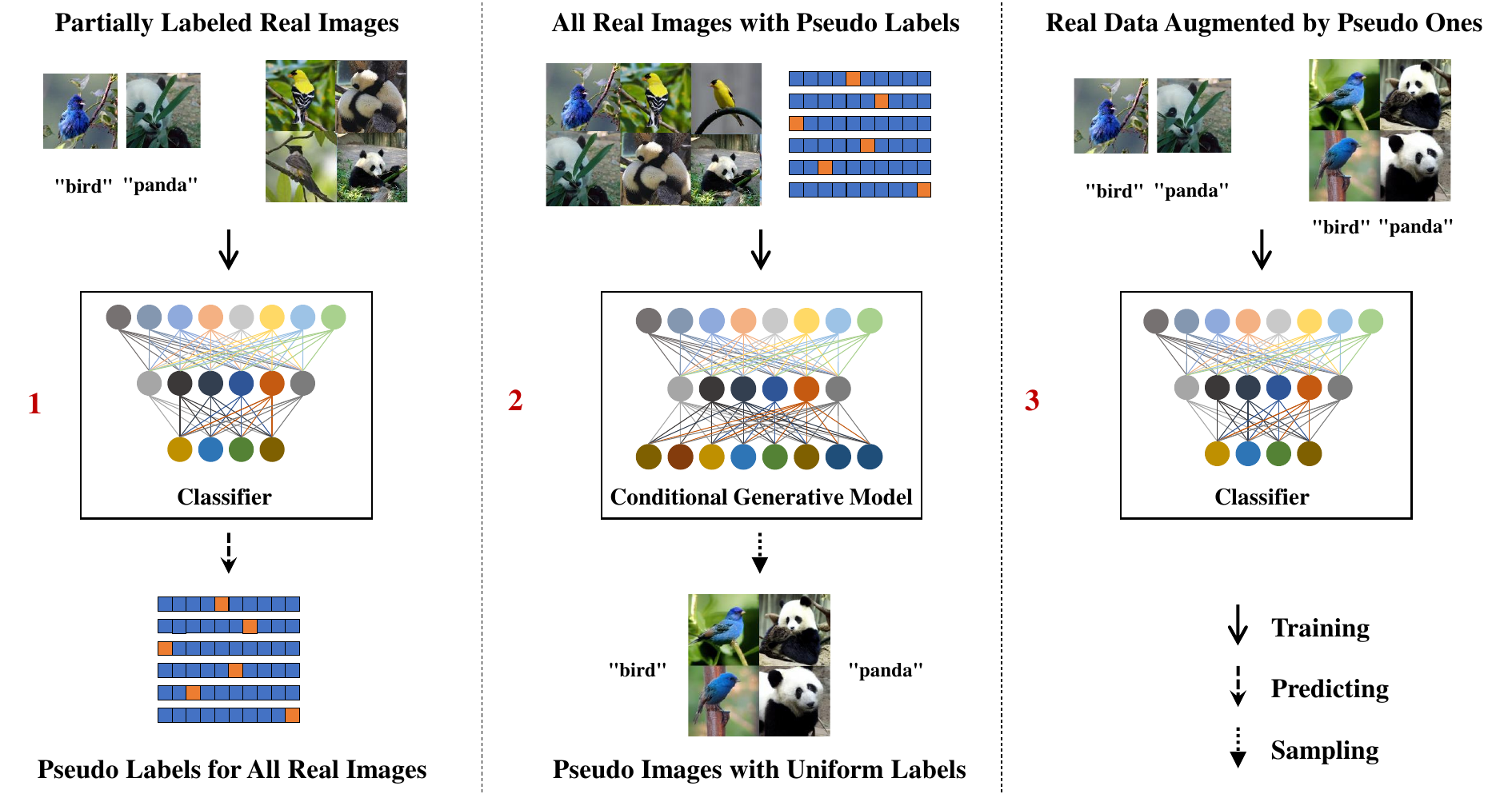}
    \caption{\textbf{An overview of DPT}. First, a (semi-supervised) classifier is trained on partially labeled data and used to predict pseudo-labels for all data. Second, a conditional  generative model is trained on all data with pseudo-labels and used to generate pseudo images given random labels. Finally, the classifier is trained or fine-tuned on real data augmented by pseudo images with labels.}
    \label{fig:flow_chart}
\end{figure*}

\textbf{Semi-ViT}~\citep{cai2022semi} is three-staged. First, it trains a ViT-based encoder $f_{\bm\theta}(\cdot)$ on all images in $\mathcal{X}$ via self-supervised methods such as MAE~\citep{DBLP:conf/cvpr/HeCXLDG22MAE}. Second, $f_{\bm\theta}(\cdot)$  is merely fine-tuned on $\mathcal{S}$ in a supervised manner. Let $\bm{\bar{\theta}}$ be the EMA of $\bm{\theta}$, and $\mathbf{x}_i^{u,s}$ and $\mathbf{x}_i^{u,w}$ denote the strong and weak augmentated versions of $\mathbf{x}_i^u$ respectively. Finally, Semi-ViT optimizes a weighted sum of two cross-entropy losses: 
\begin{flalign}
\label{eq:semi-vit}
\begin{split}
\mathcal{L} = & \ \mathcal{L}_l + \mu \mathcal{L}_u 
=  \frac{1}{N}\sum_{j=1}^N\textrm{CE}(f_{\bm\theta}(\mathbf{x}_j^l),\textrm{vec}(y_j^l)) \ + \\
&   \frac{\mu}{M}\sum_{i=1}^M \mathbb{I}[ f_{\bm{\bar{\theta}}}(\mathbf{x}_i^{u,w})_{\hat{y}_i} \geq \tau] \textrm{CE}(f_{\bm\theta}(\mathbf{x}_i^{u,s}),\textrm{vec}(\hat{y}_i)),
 \end{split}
\end{flalign}
where $f_{\bm{\bar{\theta}}}(\mathbf{x})_y$ is the logit of $f_{\bm{\bar{\theta}}}(\mathbf{x})$ indexed by $y$,  $\hat{y}_i = \arg\max_y f_{\bm{\bar{\theta}}}(\mathbf{x}_i^{u,w})_y$ is the pseudo-label, $\textrm{vec}(\cdot)$ returns the one-hot representation, and $\tau$ and $\mu$ are hyper-parameters.

\subsection{Conditional Diffusion Probabilistic Models}
\label{sec:ddpm}

\textbf{Denoising Diffusion Probabilistic Model (DDPM)}~\citep{ho2020denoising} gradually adds noise $\bm{\epsilon} \sim \mathcal{N}(\mathbf{0},\mathbf{I})$ to data $\mathbf{x}_0$ from time $t=0$ to $t=T$ in the forward process, and progressively removes noise to recover data starting at $\mathbf{x}_T \sim \mathcal{N}(\mathbf{0},\mathbf{I})$ in the reverse process. It trains a predictor $\bm{\epsilon_\theta}$ to predict the noise $\bm\epsilon$ by the following objective:
\begin{align}
\mathcal{L} = \mathbb{E}_{t,\mathbf{x}_0,\bm\epsilon}[||\bm{\epsilon_\theta}(\mathbf{x}_t,\mathbf{c},t) - \bm\epsilon||_2^2],
\end{align}
where $\mathbf{c}$ indicates conditions such as classes and texts.

\textbf{Classifier-Free Guidance (CFG)}~\citep{ho2022classifier} leverages a conditional noise predictor $\bm{\epsilon_\theta}(\mathbf{x}_t,\mathbf{c},t)$ and an unconditional noise predictor $\bm{\epsilon_\theta}(\mathbf{x}_t,t)$ in inference to improve sample quality and enhance semantics. Formally, CFG iterates the following equation starting at $\mathbf{x}_T$:
\begin{align}
    \mathbf{x}_{t-1} = \frac{1}{\sqrt{\alpha_t}}(\mathbf{x}_t - \frac{\beta_t}{\sqrt{1 - \bar{\alpha}_t}} \tilde{\bm\epsilon}_t) + \sigma_t^2 \mathbf{z},
\end{align}
where $\tilde{\bm\epsilon}_t = (1 + \omega)\bm{\epsilon_\theta}(\mathbf{x}_t,\mathbf{c},t) - \omega \bm{\epsilon_\theta}(\mathbf{x}_t,t) $, $\omega$ is the guidance strength, $\mathbf{z} \sim \mathcal{N}(\mathbf{0},\mathbf{I})$, and $\alpha_t$, $\beta_t$, $\bar{\alpha}_t$ and $\sigma_t$ are constants w.r.t. the time $t$.

\textbf{U-ViT}~\citep{bao2022all} is a ViT-based backbone for diffusion probabilistic models, which achieves excellent performance in conditional sampling on large-scale datasets.

\section{Method}
\label{sec:method}


we propose a three-stage strategy called \emph{dual pseudo training (DPT)} to advance semi-supervised generation and classification tasks, illustrated in Fig.~\ref{fig:flow_chart} and detailed as follows.

\subsection{First Stage: Train Classifier}
DPT trains a semi-supervised classifier on partially labeled data $\mathcal{S} \cup \mathcal{D}$, predicts a pseudo-label $\hat y$ of any image $\mathbf{x} \in \mathcal{X}$ by the classifier, and constructs a dataset consisting of all images with pseudo-labels, i.e. $\mathcal{S}_1 =\{(\mathbf{x}, \hat{y}) | \mathbf{x} \in \mathcal{X}\}$\footnote{For simplicity, we also use pseudo-labels instead of the ground truth for real labeled data, which are rare and have a small zero-one training loss, making no significant difference.}.
Notably, here we treat the classifier as a black box without modifying the training strategy or any hyperparameter. Therefore, any well-trained classifier can be adopted in DPT in a plug-and-play manner. Indeed, we use recent advances in self-supervised based learners for semi-supervised learning, i.e. MSN~\citep{assran2022masked}, and Semi-ViT~\citep{cai2022semi}. These two classifiers both provide the generative model with accurate, low-noise labels of high quality. 

\subsection{Second Stage: Classifier Benefits Generative Model} DPT trains a conditional generative model on all real images with pseudo-labels $\mathcal{S}_1$, samples $K$ pseudo images for any class label $y$ after training, and constructs a dataset consisting of pseudo images with uniform labels\footnote{The prior distribution of $y$ can be estimated on $\mathcal{S}$.}. We denote the dataset as $\mathcal{S}_2 = \cup_{y\in\mathcal{Y}}\{(\hat{\mathbf{x}}_{i, y}, y)\}_{i=1}^K$, where $\hat{\mathbf{x}}_{i, y}$ is the $i$-th pseudo image for class $y$. Similarly to the classifier, DPT also treats the conditional generative model as a black box. Inspired by the impressive image generation results of diffusion probabilistic  models, we take a U-ViT-based~\citep{bao2022all} denoise diffusion probabilistic model~\citep{ho2020denoising} with classifier-free guidance~\citep{ho2022classifier}
as the conditional generative model. Everything remains the same as the original work (see Sec.~\ref{sec:ddpm}) except that the set of all real images with pseudo-labels $\mathcal{S}_1$ is used for training.

We emphasize that $\mathcal{S}_1$ obtained by the first stage is necessary. In fact, $\mathcal{S}$ is of small size (e.g., one label per class) and not sufficient to train conditional diffusion models. Besides, it is unclear how to leverage unlabeled data to train such models. Built upon efficient and strong semi-supervised approaches~\citep{assran2022masked, cai2022semi}, $\mathcal{S}_1$ provides useful learning signals (with relatively small noise) to train conditional diffusion models. We present quantitative and qualitative empirical evidence in Fig.~\ref{fig:accuracy} (a) and Fig.~\ref{fig:gen_summary} respectively to affirmatively answer the first key question, namely, diffusion models with a few labels (e.g., $<0.1\%$) can generate realistic and semantically accurate images.

\subsection{Third Stage: Generative Model Benefits Classifier} 
\textbf{MSN based DPT.} We train the classifier employed in the first stage on real data augmented by $\mathcal{S}_2$ to boost classification performance. For simplicity and efficiency, we freeze the  models pre-trained by Eq.~(\ref{eq:msn}) in the first stage and replace $\mathcal{S}$ with  $\mathcal{S} \cup \mathcal{S}_2$ to train a linear probe in MSN~\citep{assran2022masked}. DPT substantially boosts the classification performance as presented in Fig.~\ref{fig:accuracy} (b). 

\textbf{Semi-ViT based DPT.} 
We freeze the models, which are pre-trained in a self-supervised manner in the first stage of Semi-ViT, and replace $\mathcal{S}$ with  $\mathcal{S} \cup \mathcal{S}_2$ to train a classifier in the third stage of Semi-ViT. We argue that pseudo images can be used in different stages of Semi-ViT and can both boost the classification performance. (see Appendix~\ref{app:different_settings_semi_vit}).

Both consistent improvements provide a positive answer to the second key question, namely, generative augmentation remains a useful approach for semi-supervised classification. Besides, we can leverage the classifier in the third stage to refine the pseudo-labels and train the generative model with one more stage. Although we observe an improvement empirically (see results in Appendix~\ref{app:more_stages}), we focus on the three-stage strategy in the main paper for simplicity and efficiency.

\section{Related Work}
\label{sec:related work}

\textbf{Semi-Supervised Classification and Generation.} The two tasks are often studied independently. For semi-supervised classification, classical work includes generative approaches based on VAE~\citep{kingma2014semi,maaloe2016auxiliary,li2017max} and GAN~\citep{DBLP:journals/corr/Springenberg15,salimans2016improved,dai2017good,gan2017triangle,li2021triple,zhang2021datasetgan}, 
and discriminative approaches with confidence regularization~\citep{joachims1999transductive,yves2004semi,lee2013pseudo,iscen2019label}, consistency regularization~\citep{miyato2015distributional,tarvainen2017mean,laine2016temporal,luo2018smooth,athiwaratkun2018there,oliver2018realistic,berthelot2019mixmatch,berthelot2019remixmatch,li2021comatch,zheng2022simmatch,sohn2020fixmatch,zhang2021flexmatch,assran2021semi,tang2022stochastic} and other approaches~\citep{wang2022unsupervised,chen2022debiased,tang2022towards,lim2021class}. Recently, large-scale self-supervised based approaches~\citep{chen2020big,grill2020bootstrap,cai2021exponential,assran2022masked,cai2022semi} have made remarkable progress in semi-supervised learning. Besides, semi-supervised conditional image generation is challenging because generative modeling is more complex than prediction. In addition, it is highly nontrivial to design proper regularization when the input label is missing. Existing work is based on VAE~\citep{kingma2014semi} or GAN ~\citep{li2017triple,luvcic2019high}, which are limited to low-resolution data (i.e., $\le$ 128 $\times$ 128) and require 10\% labels or so to achieve comparable results to supervised baselines.

In comparison, DPT handles both classification and generation tasks in extreme settings with very few labels (e.g., one label per class, $<0.1\%$ labels). Built upon recent advances in semi-supervised learners and diffusion models, DPT substantially improves the state-of-the-art results in both tasks.

\textbf{Pseudo Data and Labels.} We mention additional empirical work on generating pseudo data for supervised learning~\citep{azizi2023synthetic}, adversarial robust learning~\citep{rebuffi2021fixing,wang2023better}, 
contrastive representation learning ~\citep{jahanian2021generative} and zero-shot learning~\citep{he2022synthetic, besnier2020dataset}. Regarding theory, in the context of supervised classification,~\citet{zheng2023toward} have mentioned that when the training dataset size is small, generative data augmentation can improve the learning guarantee at a constant level. This finding can be extended to semi-supervised classification, which is left as future work.

Besides, prior work
\citep{noroozi2020self,liu2020diverse,casanova2021instance,bao2022conditional} uses cluster index or instance index as pseudo-labels to improve unsupervised generation results, which are not directly comparable to DPT. With additional few labels, DPT can generate images of much higher quality and directly control the semantics of images with class labels.

\textbf{Diffusion Models.} Recently, diffusion probabilistic models~\citep{DBLP:journals/corr/Springenberg15,ho2020denoising,song2020score,Karras2022edm} achieve remarkable progress in image generation~\citep{dhariwal2021diffusion,rombach2022high,ho2022cascaded,bao2022conditional,ho2022classifier, peebles2022scalable}, text-to-image generation~\citep{nichol2021glide,ramesh2022hierarchical,gu2022vector,rombach2022high,balaji2022ediffi}, 3D scene generation~\citep{poole2022dreamfusion}, image-editing~\citep{meng2021sdedit,choi2021ilvr,zhao2022egsde}, molecular design~\citep{hoogeboom2022equivariant,bao2022equivariant}, and semi-supervised medical science~\citep{alshenoudy2023semi, gong2023diffusion}.
There are learning-free methods~\citep{song2020denoising,DBLP:conf/iclr/BaoLZZ22,lu2022dpm++,zhang2022fast,lu2022dpm} and learning-based ones~\citep{pmlr-v162-bao22d,salimans2022progressive} 
to speed up the sampling process of diffusion models. In particular, we adopt third-order DPM-solver~\citep{lu2022dpm}, which is a recent learning-free method, for fast sampling. As for the architecture, most diffusion models rely on variants of the U-Net architecture introduced in score-based models~\citep{song2019generative} while recent work~\citep{bao2022all} proposes a promising vision transformer for diffusion models, as employed in DPT.

To the best of our knowledge, there has been little research on semi-supervised conditional diffusion models and diffusion-based semi-supervised classification, which are the focus of this paper.

\section{Experiment}
\label{sec:experiment}
We present the main experimental settings in Sec.~\ref{sec:settings}. For more details, please refer to Appendix~\ref{app:model_arc}. To evaluate the performance of DPT, we compare it with state-of-the-art conditional diffusion models and semi-supervised learners in Sec.~\ref{sec:generation_results} and Sec.~\ref{sec:classication_results} respectively. We also visualize and analyze the interaction between the stages to explain the excellent performance of DPT (see Appendix~\ref{app:analysis_generation_help_classification},~\ref{app:class_wise_analysis_classification_help_generation}). 

\subsection{Experimental Settings}
\label{sec:settings} 

\textbf{Dataset.} 
We evaluate DPT on the ImageNet~\citep{imagenet} dataset, which consists of 1,281,167 training and 50,000 validation images. In the first and third stages, we use the same pre-processing protocol for real images as the baselines~\citep{assran2022masked, cai2022semi}. For instance, in MSN, the real data are resized to $256\times256$ and then center-cropped to 224 $\times$ 224. In the second stage, real images are center-cropped to the target resolution following ~\citep{bao2022all}. In the third stage, we consider pseudo images at resolution $256\times256$ and center-crop them to 224 $\times$ 224. For semi-supervised classification, we consider the challenging settings with one, two, five labels per class and 1\% labels. The labeled and unlabeled data split is the same as that of corresponding methods~\citep{assran2022masked, cai2022semi}. We also evaluate DPT on CIFAR-10 (see detailed experiments in Appendix~\ref{app:cifar10_results}).

\textbf{Baselines.} For semi-supervised classification, we consider state-of-the-art semi-supervised approaches ~\citep{assran2022masked,cai2022semi} in the setting of low-shot (e.g., one, two, five labels per class and 1$\%$ labels) as baselines. For conditional generation, we consider the state-of-the-art diffusion models with a U-ViT architecture~\citep{bao2022all} as the baseline. 

\textbf{Model Architectures and Hyperparameters.} For a fair comparison, we use the exact same architectures and hyperparameters as the baselines~\citep{assran2022masked,cai2022semi, bao2022all}. In particular, for MSN based DPT, we use a ViT B/4 (or a ViT L/7) model~\citep{assran2022masked} for classification and a U-ViT-Large (or a U-ViT-Huge) model~\citep{bao2022all} for conditional generation. As for Semi-ViT based DPT, we use a ViT-Huge model~\cite{cai2022semi} for classification and a U-ViT-Huge model~\citep{bao2022all} for conditional generation. More details are provided in Appendix~\ref{app:model_arc} for reference.

\textbf{Evaluation metrics.}
We use the top-1 accuracy on the validation set to evaluate classification performance. For a comprehensive evaluation of generation performance, we first consider the Fréchet inception distance (FID)~\citep{heusel2017gans}, sFID~\citep{nash2021generating}, Inception Score (IS)~\citep{salimans2016improved}, precision, and recall~\citep{kynkaanniemi2019improved} on 50K generated samples. We calculate all generation metrics based on the implementation of ADM~\citep{dhariwal2021diffusion}. We also add the metric $\text{FID}_{\text{CLIP}}$, which operates similarly to FID but substitutes the Inception-V3 feature spaces with CLIP features, to eliminate confusion that FID can be artificially reduced by aligning the histograms of Top-N classifications without the actual improvement of image quality~\citep{kynkaanniemi2022role}.


\textbf{Implementation.} DPT is easy to understand and implement. In particular, it only requires several lines of code based on the implementation of the classifier and conditional diffusion model. We provide the pseudocode of DPT in the style of PyTorch in Appendix~\ref{app:pseudocode}.

\textbf{The choice of \emph{K} and \emph{CFG}.} 
We conduct detailed ablation experiments on the number of augmented pseudo images per class (i.e., \emph{K}) and the classifier-free guidance scale (i.e., \emph{CFG}) in Appendix~\ref{app:ablation_study} and find that the optimal $K$ value is 128 and the optimal $CFG$ values for different ImageNet resolutions are $0.8$ for $128\times128$, $0.4$ for $256\times256$, and $0.7$ for $512\times512$.

\textbf{The choice of resolution and number of labels.} We were primarily driven by the task of
ImageNet 256×256 generation to systematically compare with a large family of baselines. In this
context, we conducted detailed experiments, including settings with one, two, five labels per
class, and 1\% labels. We find that the performance of DPT with five labels per class is
comparable to the supervised baseline, leading us to use this setting as the default in our other
tasks such as ImageNet 128×128 and ImageNet 512×512 generation.

\begin{table}
\caption{\label{tab:sota_more_generation_128}  \textbf{Image generation results on ImageNet $\mathbf{128\bm{\times}128}$.} $^\dagger$ labels the results taken from the corresponding references and $^\star$ labels baseline achieved by us. We \textbf{bold} the best result under the corresponding setting. \emph{With $<$ 0.1\% labels, DPT outperforms strong semi-supervised generative models S$^3$GAN~\citep{luvcic2019high}.}}
\small
\centering
\vskip 0.15in
\begin{tabular}{lllccccc} 
\toprule
Method & Model & Label fraction & FID-50K $\downarrow$ & IS $\uparrow$ 
\mbox{} \vspace{.03cm} \\
& & ($\#$ labels/class) & & \\
\midrule
U-ViT-Huge(\textbf{supervised baseline})$^\star$ & Diff. & 100$\%$ & 4.53 & 219.8 \\
\midrule
S$^3$GAN~\citep{luvcic2019high}$^\dagger$ & GAN & 5$\%$ & 10.4 & 59.6 \\
S$^3$GAN~\citep{luvcic2019high}$^\dagger$ & GAN & 10$\%$ & 8.0 & 78.7 \\
S$^3$GAN~\citep{luvcic2019high}$^\dagger$ & GAN & 20$\%$ & 7.7 & 83.1 \\
DPT (\textbf{ours}, with U-ViT-Huge and MSN) & Diff. & $<0.1\% (1)$ & 4.59 & 153.6 \\
DPT (\textbf{ours}, with U-ViT-Huge and MSN) & Diff. & $<0.4\% (5)$ & \textbf{4.58} & \textbf{210.9} \\
\bottomrule
\end{tabular}
\end{table}

\begin{table}[t!]
\caption{\label{tab:sota_more_generation}  \textbf{Image generation results on ImageNet $\mathbf{256\bm{\times}256}$.} $^\dagger$ labels the results taken from the corresponding references and $^\star$ labels baselines achieved by us. DPT and the corresponding baselines employ the same model architectures~\citep{bao2022all}. \emph{With $<$ 0.4\% labels, DPT outperforms strong conditional generative models with full labels, including CDM~\citep{ho2022cascaded}, ADM~\citep{dhariwal2021diffusion} and LDM~\citep{rombach2022high}.} We \textbf{bold} the best result achieved with full labels and \underline{underline} the best result achieved with few labels. For a fair comparison, we also list the parameters of the diffusion model, including its auxiliary components.}
\small
\centering
\vskip 0.15in
\resizebox{\linewidth}{!}{
\begin{tabular}{lllccccccc} 
\toprule
Method & Model & Label fraction & FID $\downarrow$ & $\text{FID}_{\text{CLIP}}$ $\downarrow$ & sFID $\downarrow$ & IS $\uparrow$ & Precision $\uparrow$ & Recall $\uparrow$ & $\#$ Params 
\mbox{} \vspace{.03cm} \\
& & ($\#$ labels/class) & & \\
\midrule
IC-GAN~\citep{casanova2021instance}$^\dagger$ & GAN & 0$\%$ & 15.6 & - & 59.0 & - & - & - & - \\
\midrule
BigGAN-deep~\citep{brock2018large}$^\dagger$ & GAN & 100$\%$ & 6.95 & - & 7.36 & 171.4 & \underline{0.87} & 0.28 & - \\
StyleGAN-XL~\citep{sauer2022stylegan}$^\dagger$ & GAN & 100$\%$ & 2.30 & - & \textbf{4.02} & \underline{265.12} & 0.78 & 0.53 & - \\
\midrule
IDDPM~\citep{nichol2021improved}$^\dagger$ & Diff. & 100$\%$ & 12.26 & - & 5.42 & - & 0.70 & \textbf{0.62} & 550M \\
CDM~\citep{ho2022cascaded}$^\dagger$ & Diff. & 100$\%$ & 4.88 & - & - & 158.71 & - & - & - \\
ADM~\citep{dhariwal2021diffusion}$^\dagger$ & Diff. & 100$\%$ & 3.94 & - & 6.14 & 215.84 & 0.83 & 0.53 & 673M \\
LDM-4-G~\citep{rombach2022high}$^\dagger$ & Diff. & 100$\%$ & 3.60 & - & - & 247.67 & \textbf{0.87} & 0.48 & 455M \\
DiT-XL/2-G~\citep{peebles2022scalable} $^\dagger$ & Diff. &  100$\%$ &  \textbf{2.27} & - &  \underline{4.60} & \textbf{278.24} & 0.83 & 0.57 & 675M \\
U-ViT-Large~\citep{bao2022all}$^\dagger$ & Diff. & 100$\%$ & 3.40 & - & 6.63 & 219.94 & 0.83 & 0.52 & 371M \\
\midrule
\emph{With U-ViT-Large} \\
~~~~\textbf{Supervised baseline}$^\star$ & Diff. & 100$\%$ & 3.31 & 2.39 & 6.68 & 221.61 & 0.83 & 0.53 & 371M  \\
~~~~\textbf{Unsupervised baseline}$^\star$
 & Diff. & 0$\%$ & 27.99 & 5.40 & 7.03 & 33.86 & 0.60 & \underline{0.62} & 371M \\
~~~~DPT (\textbf{ours}, with MSN) & Diff. & $ < 0.1\% 
 (1)$ & 4.34  & 2.57 & 6.68 & 162.96 & 0.80 & 0.53 & 371M \\
~~~~DPT (\textbf{ours}, with MSN) & Diff. & $<0.2\%  (2)$ & 3.44 & 2.37 & 6.58 & 199.74 & 0.82 & 0.53 & 371M \\
~~~~DPT (\textbf{ours}, with MSN) & Diff. & $<0.4\% (5)$ & 3.37 & 2.35   & 6.71 & 217.53 & 0.83 & 0.52 & 371M \\
~~~~DPT (\textbf{ours}, with MSN)  & Diff. & $1\% (\approx 12)$ & 3.35 & 2.34 & 6.66 & 223.09 & 0.83 & 0.52 & 371M \\
\midrule
\emph{With U-ViT-Huge} \\
~~~~\textbf{Supervised baseline}$^\dagger$ & Diff. & 100$\%$ & \underline{2.29} & \textbf{1.75} & 5.68 & 263.88 & 0.82 & 0.57 & 585M \\
~~~~DPT (\textbf{ours}, with MSN) & Diff. & $<0.1\%  (1)$ & 3.08 & 1.84 & 5.56 & 201.68 & 0.80 & 0.58 & 585M \\
~~~~DPT (\textbf{ours}, with MSN) & Diff. & $<0.2\%  (2)$ & 2.52 & 1.81 & 5.49 & 230.34 & 0.81 & 0.57 & 585M \\
~~~~DPT (\textbf{ours}, with MSN) & Diff. & $<0.4\% (5)$ & 2.50 & 1.82 & 5.54 & 243.10 & 0.83 & 0.55 & 585M \\
~~~~DPT (\textbf{ours}, with Semi-ViT) & Diff. & $1\% (\approx 12)$ & 2.42 & \underline{1.77} & 5.48 & 259.93 & 0.82 & 0.56 & 585M \\
\bottomrule
\end{tabular}}
\end{table}

\subsection{Image Generation with Few Labels}
\label{sec:generation_results}
We show that diffusion models with a few labels can generate realistic and semantically accurate images. In particular, DPT achieves better results than semi-supervised methods on ImageNet $128 \times 128$ and comparable results to supervised methods on both ImageNet $256 \times 256$ and $512 \times 512$.


We evaluate semi-supervised generation performance of DPT on \textbf{ImageNet $\mathbf{128\bm{\times}128}$}, as shown in Tab.~\ref{tab:sota_more_generation_128}. In particular, DPT with only $<0.1\%$ labels outperforms the SOTA semi-supervised generative model S$^3$GAN~\citep{luvcic2019high} with 20$\%$ labels (FID 4.59 vs. 7.7), suggesting DPT has superior label efficiency.



In Tab.~\ref{tab:sota_more_generation}, we compare DPT with state-of-the-art generative models on \textbf{ImageNet $\mathbf{256\bm{\times}256}$}. We construct highly competitive baselines based on diffusion models with U-ViT-Large~\citep{bao2022all}. According to Tab.~\ref{tab:sota_more_generation}, our supervised and unsupervised baselines achieve an FID of 3.31 and 27.99, respectively. Leveraging the pseudo-labels predicted by the strong semi-supervised learner~\citep{assran2022masked}, DPT with few labels improves the unconditional baseline significantly and is even comparable to the supervised baseline under all metrics.  In particular, \emph{with only two labels} per class, DPT improves the FID of the unsupervised baseline by 24.55 and is comparable to the supervised baseline with a gap of 0.13. Moreover, we also construct more competitive baselines based on U-ViT-Huge to advance DPT. \emph{With one (i.e., $<$ 0.1\%) label per class}, our more powerful DPT achieves an FID of 3.08, outperforming strong supervised diffusion models including IDDPM~\citep{nichol2021improved}, CDM~\citep{ho2022cascaded}, ADM~\citep{dhariwal2021diffusion} and LDM~\citep{rombach2022high}. Additionally, with 1\% labels, DPT achieves an FID of 2.42, comparable to the state-of-the-art supervised diffusion model~\citep{peebles2022scalable}. Lastly, DPT with few labels performs comparably to the fully supervised baseline under the $\text{FID}_{\text{CLIP}}$ metric, which suggests that DPT can generate high-quality samples and does not achieve a lower FID solely due to better Top-N alignment.

We also conduct an experiment on higher resolution (i.e., $\mathbf{512\bm{\times}512}$) in Tab.~\ref{tab:sota_more_generation_512}, \emph{with five (i.e., $<$ 0.4\%) labels}, DPT achieves an FID of 4.05, which is the same as that of the supervised baseline. The above quantitative results demonstrate that DPT can achieve excellent generation performance and label efficiency at diverse resolutions. Qualitatively, as presented in Fig.~\ref{fig:gen_summary}, DPT can generate realistic, diverse, and semantically correct images even with a single label, which agrees with the quantitative results in Tab.~\ref{tab:sota_more_generation} and Tab.~\ref{tab:sota_more_generation_512}. We provide more samples and failure cases in Appendix~\ref{appen:more_samples} and a detailed class-wise analysis to show how classification helps generation in Appendix~\ref{app:class_wise_analysis_classification_help_generation}.

Besides, Tab.~\ref{tab:generation_cifar10} in Appendix~\ref{app:cifar10_results} compares DPT with state-of-the-art generative models on CIFAR-10. DPT achieves competitive performance using only $0.08\%$ 
labels with EDM~\citep{Karras2022edm}, which relies on full labels (FID 1.81 vs. 1.79). This result demonstrates the generalizability of DPT on different datasets. 


\begin{table}[t!]
\caption{\label{tab:sota_more_generation_512}  \textbf{Image generation results on ImageNet $\mathbf{512\bm{\times}512}$.} $^\dagger$ labels the results taken from the corresponding references. We \textbf{bold} the best result under the corresponding setting.}
\small
\centering
\vskip 0.15in
\begin{tabular}{lllccccc} 
\toprule
Method & Model & Label fraction & FID-50K $\downarrow$ & IS $\uparrow$ 
\mbox{} \vspace{.03cm} \\
& & ($\#$ labels/class) & & \\
\midrule
BigGAN-deep~\citep{brock2018large}$^\dagger$ & GAN & 100$\%$ & 8.43 &  177.90  \\
StyleGAN-XL~\citep{sauer2022stylegan}$^\dagger$ & GAN & 100$\%$ & \textbf{2.41} & \textbf{267.75} \\
\midrule
ADM~\citep{dhariwal2021diffusion}$^\dagger$ & Diff. &  100$\%$ &  3.85 &  221.72 \\
DiT-XL/2-G~\citep{peebles2022scalable}$^\dagger$ & Diff. &  100$\%$ &  3.04 &  240.82 \\
U-ViT-Huge (\textbf{supervised baseline})$^\dagger$ & Diff. & 100$\%$ & 4.05 & 263.79 \\
\midrule
DPT (\textbf{ours}, with U-ViT-Huge and MSN)  & Diff. & $<0.4\% (5)$ & \textbf{4.05} & \textbf{252.08} \\
\bottomrule
\end{tabular}
\end{table}

\begin{table*}
\caption{\label{tab:sota_more_semi} \textbf{Top-1 accuracy on the ImageNet validation set with few labels.} $^\dagger$ labels the results taken from corresponding references, 
$^\ddagger$ labels the results taken from~\citet{assran2022masked} and $^\star$ labels the baselines reproduced by us. DPT and the corresponding baseline employ exactly the same classifier architectures. \emph{With one, two, five labels per class and 1\% labels, DPT improves the state-of-the-art semi-supervised learner~\citep{assran2022masked, cai2022semi} consistently and substantially.} We \textbf{bold} the best result under the corresponding setting and \underline{underline} the second-best result.}
\vskip 0.15in
\small
\centering
\begin{tabular}{llccccc}
\toprule
{Method}  & {Architecture} & \multicolumn{4}{c}{Top-1 accuracy $\uparrow$ given $\#$ labels per class (label fraction)} 
\mbox{} \vspace{.08cm} \\
&  & $1 (<0.1\%)$ & $2 (<0.2\%)$ & $5 (<0.5\%)$ & $\approx12 (1\%)$ \\
\midrule
EMAN~\citep{cai2021exponential}$^\dagger$ & ResNet-50 & - & - & - & 63.0 \\
PAWS~\citep{assran2021semi}$^\dagger$ & ResNet-50 & - & - & - & 66.5 \\
BYOL~\citep{grill2020bootstrap}$^\dagger$ & ResNet-200  & - & - & - & 71.2 \\
SimCLRv2~\citep{chen2020big}$^\dagger$ & ResNet-152 & - & - & - & 76.6 \\
\midrule
Semi-ViT~\citep{cai2022semi}$^\dagger$ & ViT-Huge & - & - & - & \underline{80.0} \\
iBOT~\citep{zhou2021ibot}$^\ddagger$ & ViT-B/16 & 46.1 $\pm$ 0.3 & 56.2 $\pm$ 0.7 & 64.7 $\pm$ 0.3 & - \\
DINO~\citep{caron2021emerging}$^\ddagger$ & ViT-B/8 & 45.8 $\pm$ 0.5 & 55.9 $\pm$ 0.6 & 64.6 $\pm$ 0.2 & - \\
MAE~\citep{DBLP:conf/cvpr/HeCXLDG22MAE}$^\ddagger$ & ViT-H/14 & 11.6 $\pm$ 0.4 & 18.6 $\pm$ 0.2 & 32.8 $\pm$ 0.2 & - \\
MSN~\citep{assran2022masked}$^\dagger$ & ViT-B/4 & 54.3 $\pm$ 0.4 & 64.6 $\pm$ 0.7 & 72.4 $\pm$ 0.3 & 75.7 \\
MSN~\citep{assran2022masked}$^\dagger$ & ViT-L/7 & 57.1 $\pm$ 0.6 & 66.4 $\pm$ 0.6 &72.1 $\pm$ 0.2 & 75.1 \\
\midrule 
MSN (\textbf{baseline})$^\star$  & ViT-B/4 & 52.9 & 64.9 & 72.4 & - \\
DPT (\textbf{ours}) & ViT-B/4 & \underline{58.6}   & \textbf{69.5} & \textbf{74.4} &  - \\
\midrule
MSN (\textbf{baseline})$^\star$ & ViT-L/7 & 56.2 & 66.5 & 72.0 & - \\
DPT (\textbf{ours}) & ViT-L/7 & \textbf{58.9}  & \underline{69.2}  & \underline{73.4}  & - \\
\midrule
Semi-ViT (\textbf{baseline})$^\star$ & ViT-Huge & - & - & - & 79.4 \\
DPT (\textbf{ours}) & ViT-Huge & - & - & - & \textbf{80.2} \\
\bottomrule
\end{tabular}
\end{table*}


\subsection{Image Classification with Few Labels}
\label{sec:classication_results}
We demonstrate that generative augmentation remains a useful approach for semi-supervised classification aided by diffusion models. In particular, DPT achieves state-of-the-art semi-supervised classification performance on ImageNet datasets across various settings and the second-best results on CIFAR-10.

Tab.~\ref{tab:sota_more_semi} compares DPT with state-of-the-art semi-supervised classifiers on the ImageNet validation set with few labels. Specifically, DPT outperforms strong semi-supervised baselines ~\citep{assran2022masked,cai2022semi} consistently and substantially \emph{with one, two, five labels per class and 1$\%$ labels} and achieves state-of-the-art top-1 accuracies of 59.0, 69.5, 74.4 and 80.2, respectively. In particular, with two labels per class, DPT leverages the pseudo images generated by the diffusion model and improves MSN with ViT-B/4 by an accuracy of 4.6\%.  Besides, we compare the performance of DPT with that of SOTA fully supervised models (as shown in Tab.~\ref{tab:compare_supervised} in Appendix~\ref{app:different_settings_semi_vit}) and find that DPT performs comparably to Inception-v4 \citep{szegedy2017inception}, using only 1$\%$ labels.

Moreover, Tab.~\ref{tab:classification_cifar10} in Appendix~\ref{app:cifar10_results} compares DPT with state-of-the-art semi-supervised classifiers on CIFAR-10. DPT with four labels per class achieves the second-best error rate of 4.68{\scriptsize $\pm$0.17}\%.

\section{Conclusions}
\vspace{-.1cm}
This paper presents a simple yet effective training strategy called DPT for conditional image generation and classification in semi-supervised learning. Empirically, we demonstrate that DPT can achieve SOTA semi-supervised generation and classification performance on ImageNet datasets across various settings. DPT probably inspires future work in diffusion models and semi-supervised learning.  

\textbf{Limitation.} One limitation of DPT is directly using the pseudo images to improve the performance of DPT for its simplicity and effectiveness while we could use pre-trained models like CLIP to filter out noisy image-label pairs that images do not semantically align well with the label. Another limitation pertains to the direct use of pseudo labels. Given our use of classifier-free guidance, we have the flexibility to assign low-confidence pseudo labels to the null token with a high probability, which aids in filtering out noisy pseudo labels.

\textbf{Social impact.} We believe that DPT can benefit real-world applications with few labels (e.g., medical analysis). However, the proposed semi-supervised diffusion models may aggravate social issues such as ``DeepFakes''. The problem can be relieved by automatic detection with machine learning, which is an active research area. 

\section*{Acknowledgement}

This work was supported by NSF of China (Nos. 62076145); Beijing Outstanding Young Scientist Program (No. BJJWZYJH012019100020098); Major Innovation \& Planning Interdisciplinary Platform for the ``Double-First Class" Initiative, Renmin University of China; the Fundamental Research Funds for the Central Universities, and the Research Funds of Renmin University of China (No. 22XNKJ13). C. Li was also sponsored by Beijing Nova Program (No. 20220484044).

\bibliography{ref}
\bibliographystyle{IEEEtranN}

\newpage
\appendix
\section{Results of CIFAR-10}
\label{app:cifar10_results}
To evaluate the generalizability of DPT across different datasets, we also conduct an experiment on CIFAR-10~\citep{krizhevsky2009learning}. 
\subsection{Baselines}
For semi-supervised classification, we consider a state-of-the-art method called FreeMatch~\citep{wang2022freematch} as the baseline. For conditional generation, we consider a state-of-the-art method called EDM~\citep{Karras2022edm} as the baseline. 
The training configuration of EDM is variance preserving (VP) ~\citep{song2020score}, as it achieves slightly better generation performance compared to the alternative configuration of variance exploding (VE)~\citep{song2020score}.

\subsection{Settings}
In the second stage of DPT, we generate pseudo images using the same sampling process as EDM~\citep{Karras2022edm}. We set the number of augmented pseudo images per class, i.e., \emph{K}, to 1001 as the default value if not specified. In the third stage of DPT, we replace $\mathcal{S}$ with  $\mathcal{S} \cup \mathcal{S}_2$ to re-train FreeMatch~\citep{wang2022freematch}.
\subsection{Evaluation metrics}
We use the error rate on the validation set to evaluate classification performance and consider the Fréchet inception distance (FID) score ~\citep{heusel2017gans} to evaluate generation performance.
\subsection{Image Generation with Few Labels}
Tab.~\ref{tab:generation_cifar10} presents a quantitative comparison of DPT with state-of-the-art generative models on the CIFAR-10 generation benchmark. In particular, DPT achieves an FID of 1.81 with only \emph{four (i.e., $0.08\%$) labels per class}, outperforming strong supervised generative models such as StyleGAN-XL~\citep{sauer2022stylegan} and IDDPM~\citep{nichol2021improved}, and even demonstrating competitive performance compared to the state-of-the-art supervised generative model EDM~\citep{Karras2022edm}.

\begin{table}[!htb]
\caption{\label{tab:generation_cifar10}  \textbf{Image generation results on CIFAR-10 32$\times$32.}}
\small
\centering
\vskip 0.15in
\begin{tabular}{lllc} 
\toprule
Method & Model & Label fraction & FID-50K $\downarrow$ 
\mbox{} \vspace{.03cm} \\
& & ($\#$ labels/class) & \\
\midrule
StyleGAN2-ADA~\citep{karras2020training} & GAN & 100$\%$ & 2.92 \\
StyleGAN-XL\citep{sauer2022stylegan} & GAN & 100$\%$ & 1.85 \\
\midrule
EDM~\citep{Karras2022edm} & Diff. & 0$\%$ & 1.97 \\
DDPM~\citep{ho2020denoising} & Diff. & 100$\%$ & 3.17 \\
IDDPM~\citep{nichol2021improved} & Diff. & 100$\%$ & 2.90 \\
U-ViT~\citep{bao2022all} & Diff. & 100$\%$ & 3.11 \\
EDM~\citep{Karras2022edm} & Diff. & 100$\%$ & \textbf{1.79} \\
\midrule
DPT (\textbf{ours}, with EDM and FreeMatch)  & Diff. & $0.08\%$ (4) & \textbf{1.81} \\
\bottomrule
\end{tabular}
\end{table}

\subsection{Image Classification with Few Labels}
Tab.~\ref{tab:classification_cifar10} presents a comparison of DPT with state-of-the-art semi-supervised classifiers on CIFAR-10. DPT outperforms competitive  baselines~\citep{wang2022freematch, zhang2021flexmatch, sohn2020fixmatch} substantially with four labels per class, achieving the second-best error rate of 4.68{\scriptsize $\pm$0.17}\%. Meanwhile, it's worth noting that the state-of-the-art method Full-flex and our work DPT are orthogonal. Since DPT is a flexible framework, integrating FullFlex~\citep{chen2023boosting} into DPT could potentially lead to further performance improvements.

\begin{table*}
\caption{\label{tab:classification_cifar10}  \textbf{Error rates on CIFAR-10 32$\times$32.} \textbf{Bold} indicates the best result and \underline{underline} indicates the second-best result. $^\dagger$ labels the results taken from corresponding references, and $^\star$ labels the baselines reproduced by us.}
\vskip 0.15in
\small
\centering
\begin{tabular}{lcc}
\toprule
{Method}  & \multicolumn{2}{c}{Error rate $\downarrow$ } \\
given $\#$ labels per class (label fraction) &  4 (0.08\%) & 25 (0.5\%) \\
\midrule
$\Pi$ Model~\citep{rasmus2015semi}$^\dagger$ & 74.34{\scriptsize $\pm$1.76} & 46.24{\scriptsize $\pm$1.29} \\
Pseudo Label~\citep{lee2013pseudo}$^\dagger$ & 74.61{\scriptsize $\pm$0.26}  & 46.49{\scriptsize $\pm$2.20} \\
VAT ~\citep{miyato2018virtual}$^\dagger$ & 74.66{\scriptsize $\pm$2.12} & 41.03{\scriptsize $\pm$1.79} \\
MeanTeacher \citep{tarvainen2017mean}$^\dagger$ & 70.09{\scriptsize $\pm$1.60} & 37.46{\scriptsize $\pm$3.30} \\
MixMatch \citep{berthelot2019mixmatch}$^\dagger$ & 36.19{\scriptsize $\pm$6.48} & 13.63{\scriptsize $\pm$0.59} \\
ReMixMatch \citep{berthelot2019remixmatch}$^\dagger$ & 9.88{\scriptsize $\pm$1.03} & 6.30{\scriptsize $\pm$0.05} \\
UDA \citep{xie2020unsupervised}$^\dagger$ & 10.62{\scriptsize $\pm$3.75} & 5.16{\scriptsize $\pm$0.06} \\
FixMatch \citep{sohn2020fixmatch}$^\dagger$ & 7.47{\scriptsize $\pm$0.28} & 4.86{\scriptsize $\pm$0.05} \\
PPF~\citep{tang2022towards} $^\dagger$ &  7.71{\scriptsize $\pm$3.06} & 4.84{\scriptsize $\pm$0.17} \\
STOCO~\citep{tang2022stochastic}$^\dagger$ & 7.18{\scriptsize $\pm$1.95} & 4.78{\scriptsize $\pm$0.30}\\
Dash \citep{xu2021dash}$^\dagger$ & 8.93{\scriptsize $\pm$3.11} & 5.16{\scriptsize $\pm$0.23} \\
MPL \citep{pham2021meta}$^\dagger$ & 6.62{\scriptsize $\pm$0.91} & 5.76{\scriptsize $\pm$0.24} \\
FlexMatch \citep{zhang2021flexmatch}$^\dagger$ & 4.97{\scriptsize $\pm$0.06} & 4.98{\scriptsize $\pm$0.09} \\
DST~\citep{chen2022debiased}$^\dagger$ & 5.00 & - \\
FullFlex \citep{chen2023boosting}$^\dagger$ & \textbf{4.44}{\scriptsize $\pm$0.15} & \textbf{4.39}{\scriptsize $\pm$0.04} \\
\midrule
FreeMatch \citep{wang2022freematch}$^\dagger$ & 4.90{\scriptsize $\pm$0.04} & \underline{4.88}{\scriptsize $\pm$0.18} \\
FreeMatch (\textbf{baseline})$^\star$ & 4.93{\scriptsize $\pm$0.13} & - \\
DPT (\textbf{ours}) with EDM and FreeMatch & \underline{4.68}{\scriptsize $\pm$0.17} & - \\
\bottomrule
\end{tabular}
\end{table*}


\section{Pseudocode of DPT}
\label{app:pseudocode}
Algorithm~\ref{alg:dpt} presents the pseudocode of DPT in the PyTorch style. Based on the implementation of the classifier and the conditional generative model, DPT is easy to implement with a few lines of code in PyTorch. 

\begin{algorithm}[t!]
   \caption{Pseudocode of DPT in a PyTorch style.}
   \label{alg:dpt}
   
    \definecolor{codeblue}{rgb}{0.25,0.5,0.5}
    \lstset{
      basicstyle=\fontsize{7.2pt}{7.2pt}\ttfamily\bfseries,
      commentstyle=\fontsize{7.2pt}{7.2pt}\color{codeblue},
      keywordstyle=\fontsize{7.2pt}{7.2pt},
    }
\begin{lstlisting}[language=python]
# Classifier: a classifier
# Generative_model: conditional generative models, such as diffusion models
# real_labeled_data: real labeled data
# real_unlabeled_data: real unlabeled data
# all_real_images: all images in real labeled and unlabeled data
# C: the number of classes in real labeled and unlabeled data
# K: the number of pseudo samples
# Uniform: uniform sampling function

### first stage: 

# train a classifier
Classifier.train([(real_labeled_data.images, real_labeled_data.labels),             
                  (real_unlabeled_data.images, )]) 

# predict pseudo labels for all real images 
pseudo_labels = Classifier.predict(all_real_images) 

### second stage

# train a conditional diffusion model
Generative_model.train([(all_real_images, pseudo_labels)]) 

uniform_labels = Uniform(C, K) # uniformly sample K labels from [0, C)

# sample K pseudo images by Generative_model
pseudo_images = Generative_model.sample(uniform_labels) 

### third stage

# re-train the classifier 
Classifier.train([(real_labeled_data.images,real_labeled_data.labels),
                  (pseudo_images, uniform_labels), 
                  (real_unlabeled_data.images, )]) 

\end{lstlisting}
\end{algorithm}

\begin{table}[t!]
\caption{\label{tab:code_used_and_license} \textbf{The code links and licenses.}}
\vskip 0.15in
\small
\centering
\begin{tabular}{lccccc}
\toprule
Method & Link  &  License \\
\midrule
ADM & \url{https://github.com/openai/guided-diffusion} &  MIT License \\
LDM & \url{https://github.com/CompVis/latent-diffusion} & MIT License \\
U-ViT & \url{https://github.com/baofff/U-ViT} &  MIT License \\
DPM-Solver & \url{https://github.com/LuChengTHU/dpm-solver} & MIT License \\
FreeMatch & \url{https://github.com/TorchSSL/TorchSSL} & MIT License \\
Semi-ViT & \url{https://github.com/amazon-science/semi-vit} & Apache License \\
MSN & \url{https://github.com/facebookresearch/msn} &  CC BY-NC 4.0 \\
EDM & \url{https://github.com/NVlabs/edm} & CC BY-NC-SA 4.0 \\

\bottomrule
\end{tabular}
\end{table}

\begin{table}[t!]
\caption{\label{tab:model_architecture} \textbf{Model architectures in semi-supervised classifier and U-ViT.}}
\vskip 0.15in
\small
\centering
\begin{tabular}{lccccc}
\toprule
Model  & Param & \# Layers & Hidden Size & MLP Size & \# Heads \\
\midrule
\emph{Semi-Supervised Classifier (MSN)} \\ 
~~~~ViT B/4 & 86M  & 12 & 768 & 3072 & 12 \\
~~~~ViT L/7 & 304M & 24 & 1024 & 4096 & 16 \\
\emph{Semi-Supervised Classifier (Semi-ViT)} \\ 
~~~~ViT Huge & 632M  & 32 & 1280 & 5120 & 16 \\
\midrule
\emph{Conditional Diffusion Model (U-ViT)} \\
~~~~U-ViT-Large & 371M & 21 & 1024 & 4096 & 16 \\
~~~~U-ViT-Huge & 585M &  29 & 1152 & 4608 & 16 \\
\bottomrule
\end{tabular}
\end{table}

\section{Experimental Setting}
\label{app:model_arc}

We implement DPT upon the official code of LDM~\citep{rombach2022high}, DPM-Solver~\citep{lu2022dpm}, ADM~\citep{dhariwal2021diffusion}, MSN~\citep{assran2022masked}, Semi-ViT~\citep{cai2022semi}, EDM~\citep{Karras2022edm}, FreeMatch~\citep{wang2022freematch} and U-ViT~\citep{bao2022all}, whose code links and licenses are presented in Tab.~\ref{tab:code_used_and_license}. 
All the architectures and hyperparameters are the same as the corresponding baselines~\citep{bao2022all,assran2022masked,cai2022semi,wang2022freematch, Karras2022edm}. For completeness, we briefly mention important settings and refer the readers to the original paper for more details. We also report the computational cost in Appendix.~\ref{app:cost}.

\textbf{SCDM.} We extract features of ImageNet using the self-supervised method MSN~\citep{assran2022masked} and perform k-means on these features to obtain meaningful cluster indices as conditions for training U-ViT-Large. Notably, in this way, we achieve an FID of 5.19 on ImageNet $256\times256$ without labels. However, the performance is still inferior to an FID of 3.31 achieved by supervised models.

\textbf{The usage of pseudo images in the third stage.} We focus on using pseudo images at a resolution of $256\times256$ because this resolution is closest to the commonly applied $224\times224$ resolution used for ImageNet classification. 
It is worth noting that for MSN based DPT, we utilize pseudo images generated by U-ViT-Large, except in cases where the DPT employs ViT-B/4 and has five labels per class and we use pseudo images generated by U-ViT-Huge instead. This is done to explore whether the pseudo images from the more powerful generative model can provide additional benefits to the classifier. For Semi-ViT based DPT, we employ pseudo images generated by U-ViT-Huge.

\textbf{Network architectures.} We present the network architectures in Tab.~\ref{tab:model_architecture}. 

\textbf{MSN.} MSN adopts a warm-up 
strategy over the first 15 epochs of training, which linearly increases the learning rate from 0.0002 to 0.001, and then decays the learning rate to 0 following a cosine schedule. The total training epochs are 200 and 300 for the architecture of ViT L/7 and ViT B/4, separately. The batch size is 1024 for both two architectures. Actually, we reuse the two \textbf{pre-trained} models ViT L/7 and ViT B/4 provided by MSN~\citep{assran2022masked} to reduce the training cost. After extracting the features by MSN, we use the \texttt{cyanure} package~\citep{mairal2019cyanure} to train the classifier following MSN~\citep{assran2022masked}. In particular, we run logistic regression on a single CPU core based on~\texttt{cyanure}. 

\textbf{U-ViT.} U-ViT is based on the latent diffusion~\citep{rombach2022high}. Specifically, we adopt two best configurations of U-ViT: U-ViT-Large and U-ViT-Huge. U-ViT-Large trains a transformer-based conditional generative model with a batch size of 1024, a training iteration of 300k, and a learning rate of 2e-4. U-ViT-Huge uses the same learning rate and batch size as U-ViT-Large but is trained for 500k iterations.

\textbf{EDM.} We use EDM for conditional generation on CIFAR-10 dataset. EDM trains a conditional diffusion model with a batch size of 512, a training duration of 200 Mimg, and a learning rate 1e-3.


\textbf{Semi-ViT.} We consider the best configuration of Semi-ViT with 1\% labels, i.e., ViT-Huge. In the first stage, Semi-ViT uses the pre-training model of MAE. In the second stage, Semi-ViT trains a transformer-based classifier with a batch size of 128, a training epoch of 50, and a learning rate of 0.01. In the third stage, Semi-ViT trains a transformer-based classifier with a batch size of 64, a training epoch of 50, and a learning rate of 5e-3.

\textbf{FreeMatch.} FreeMatch trains a WRN-28-2 model with a batch size of 64, a training iteration of $2^{20}$, and a learning rate 0.03.

\section{Computational Cost}
\label{app:cost}
We present the detailed computational cost of MSN based DPT and Semi-ViT based DPT on ImageNet $256\times 256$ in Tab.~\ref{tab:cost_time_msn_huge} and Tab.~\ref{tab:cost_time_semivit_huge}, respectively.

As illustrated in Tab.~\ref{tab:cost_time_msn_huge}, DPT with MSN introduces an additional computation cost of approximately $\frac{\textrm{DPT (extra~cost)}}{\textrm{Classifier~+~Generator}} = \frac{50}{8640} = 0.57\%$, which is negligible. In particular, for conditional generation, the extra overhead we introduce is the cost of training the classifier. We reuse the pre-trained MSN to extract the features, and thus the training cost of the classifier can be reduced to only 30 V100-hours, which is negligible compared to the cost of the generator. For semi-supervised classification, the extra overhead we introduce is the cost of generative augmentation. The percentage of additional time cost over MSN is approximately 201.7\%, calculated as $\frac{\textrm{Generator + DPT extra cost}}{\textrm{Classifier}}=\frac{5813}{2881}=201.7\%$. Although DPT requires nearly twice the training time compared to the MSN baseline, it's still more time-efficient than other methods like Triple-GAN~\citep{li2017triple, li2021triple}, which demands at least 5 times the training time of its classifier. 


Moreover, the percentage of additional computation cost of DPT with Semi-ViT is $\frac{\textrm{DPT (extra~cost)}}{\textrm{Classifier~+~Generator}} = \frac{3886}{9664} = 40.21\%$, as shown in Tab.~\ref{tab:cost_time_semivit_huge}. Although Semi-ViT brings more accurate pseudo labels for conditional generation, it also needs more expensive training costs, creating a trade-off between label accuracy and computational expenses.

Furthermore, the computational cost of DPT on CIFAR-10 is presented in Tab.~\ref{tab:cost_time_cifar_edm}. The percentage of additional computation cost of DPT is $\frac{\textrm{DPT (extra~cost)}}{\textrm{Classifier~+~Generator}} = \frac{169}{552} = 30.62\%$.

\begin{table}[t!]
\caption{\label{tab:cost_time_msn_huge} The training time of DPT using U-ViT-H/2 and MSN with ViT-L/7 on ImageNet $256\times256$ with 5 labels per class. U-ViT-H/2 indicates that we use the U-ViT-Huge with the input patch size of $2 \times 2$. 
We present the percentage of additional computation cost of DPT in parentheses.}
\vskip 0.15in
\small
\centering

\begin{tabular}{llll}
\toprule
Model & Process & V100-hours & Cpu-hours \\
\midrule
Classifier & Self supervised pre-training & 2850 & - \\
 & Extracting features & 30 & - \\
 & Linear classification & - & 1 \\
\midrule
Generator & Generation & 5760 & - \\
\midrule
DPT (extra cost) & Sampling & 46 & - \\
 & Extracting features & 4 & - \\
 & Linear classification & - & 3 \\
\midrule
DPT & All training (sum all above) & 8690 (\textbf{0.57\%}) & 4 \\
\bottomrule
\end{tabular}
\end{table}

\begin{table}[t!]
\caption{\label{tab:cost_time_semivit_huge} The training time of DPT using Semi-ViT and U-ViT-H/2 on ImageNet $256\times256$ with 1\% labels. U-ViT-H/2 indicates that we use the U-ViT-Huge with the input patch size of $2 \times 2$. We present the percentage of additional computation cost of DPT in parentheses. }
\vskip 0.15in
\small
\centering

\begin{tabular}{lll}
\toprule
Model & Process & V100-hours \\
\midrule
Classifier & Supervised fine-tuning & 64  \\
 & Semi-supervised fine-tuning & 3840  \\
\midrule
Generator & Generation & 5760  \\
\midrule
DPT (extra cost) & Sampling & 46  \\
 & Semi-supervised fine-tuning & 3840  \\
\midrule
DPT & All training (sum all above) & 13550 (\textbf{40.21\%}) \\
\bottomrule
\end{tabular}
\end{table}

\begin{table}[t!]
\caption{\label{tab:cost_time_cifar_edm} The training time of DPT using FreeMatch and EDM on CIFAR-10 with 4 labels per class. We present the percentage of additional computation cost of DPT in parentheses.}
\vskip 0.15in
\small
\centering

\begin{tabular}{lll}
\toprule
Model & Process & V100-hours \\
\midrule
Classifier & Classification &  168 \\
\midrule
Generator & Generation & 384 \\
\midrule
DPT (extra cost) & Sampling & 1 \\
 & Classification & 168 \\
\midrule
DPT & All training (sum all above) & 721 (\textbf{30.62\%}) \\
\bottomrule
\end{tabular}
\end{table}

\section{Thought experiment}
\label{app:thought_experiment}
Classification and class-conditional generation are dual tasks that characterize opposite conditional distributions, e.g., $p(\textrm{label} | \textrm{image})$ and $p(\textrm{image} | \textrm{label})$. Learning such conditional distributions is conceptually natural given a sufficient amount of image-label pairs. Recent advances in self-supervised learning\footnote{Self-supervised methods learn representations without labels but require full labels to obtain $p(\textrm{label} | \textrm{image})$.}~\citep{DBLP:conf/cvpr/He0WXG20Moco, DBLP:conf/icml/ChenK0H20SimCLR, DBLP:conf/cvpr/ChenH21SimSiam,grill2020bootstrap,xie2021simmim,DBLP:conf/cvpr/HeCXLDG22MAE} and diffusion probabilistic  models~\citep{sohl2015deep,ho2020denoising,song2020score,dhariwal2021diffusion,bao2022all,Karras2022edm} achieve  excellent performance in the two tasks respectively. However, both learning tasks are nontrivial in semi-supervised learning, where only a small fraction of the data are labeled (see Sec.~\ref{sec:related work} for a comprehensive review).

Most previous work solves the two tasks independently in semi-supervised learning while they can benefit mutually in intuition. The idea is best illustrated by a thought experiment with infinite model capacity and zero optimization error. Let $p( \textrm{image})$ be the true marginal distribution, from which we obtain massive samples in semi-supervised learning. Suppose we have a sub-optimal conditional distribution $p_c(\textrm{label} | \textrm{image})$ characterized by a classifier, a joint distribution $p_c(\textrm{image},\textrm{label})=p_c(\textrm{label} | \textrm{image})p( \textrm{image})$ is induced by predicting pseudo-labels for unlabeled data. Meanwhile, a conditional generative model trained on sufficient pseudo data from $p_c(\textrm{image},\textrm{label})$  can induce the same joint distribution, as long as it is Fisher consistent\footnote{It means that the returned hypothesis in a sufficiently expressive class can recover the true distribution given infinite data.}~\citep{fisher1922mathematical}. Because the generative model can further leverage the real data in a complementary way to the classifier, the induced joint distribution (denoted as $p_g(\textrm{image},\textrm{label})$) is probably closer to the true distribution than $p_c(\textrm{image},\textrm{label})$. 
Similarly, the classifier can be enhanced by training on pseudo data sampled from $p_g(\textrm{image},\textrm{label})$). In conclusion, the classifier and the conditional generative model can benefit mutually through pseudo-labels and data in the ideal case.

\section{Additional Results and Discussions}


\subsection{More Samples and Failure Cases}
\label{appen:more_samples}

\begin{figure*}[t!]
\begin{center}
\subfloat[Random samples with \emph{one} label per class. \emph{Left:} ``Gondola''.  \emph{Right:} ``Yellow lady's slipper''.]{\includegraphics[width=\linewidth]{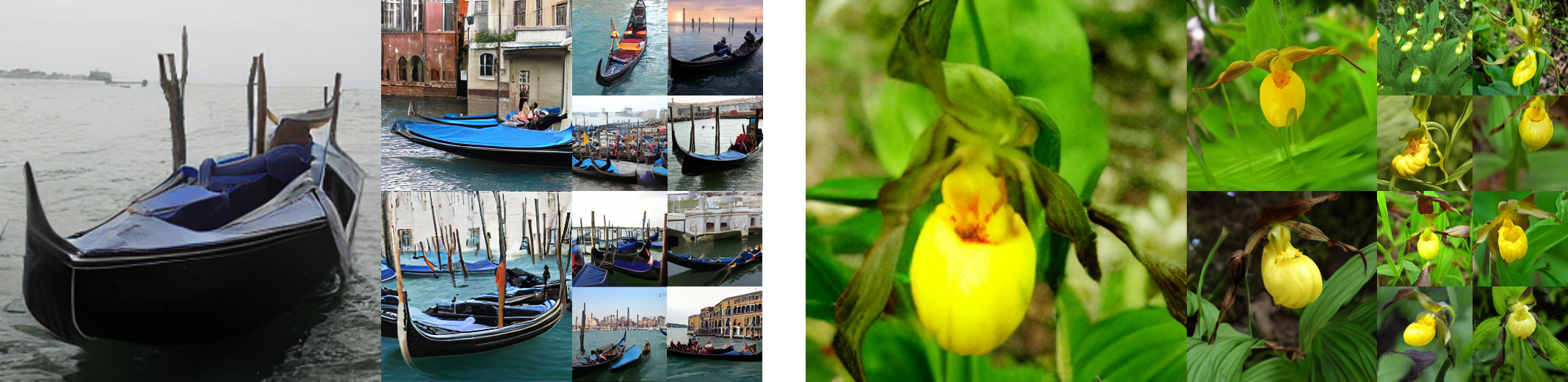}}\\
\subfloat[Random samples with \emph{two} labels per class. \emph{Left:} ``Triceratops''.  \emph{Right:} ``Echidna''.]{\includegraphics[width=\linewidth]{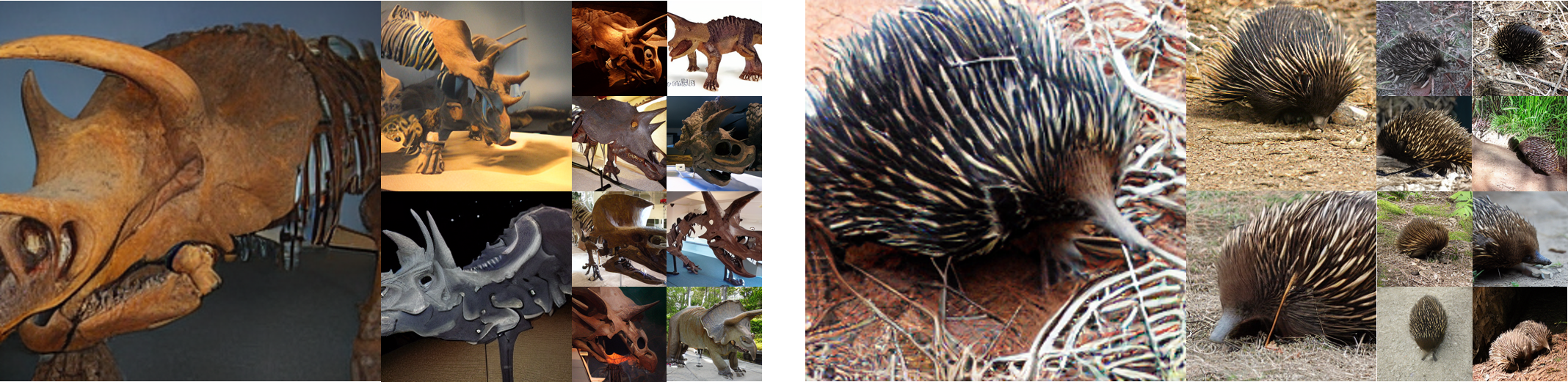}}\\
\subfloat[Random samples with \emph{five} labels per class. \emph{Left:} ``School bus''.  \emph{Right:} ``Fig''.]{\includegraphics[width=\linewidth]{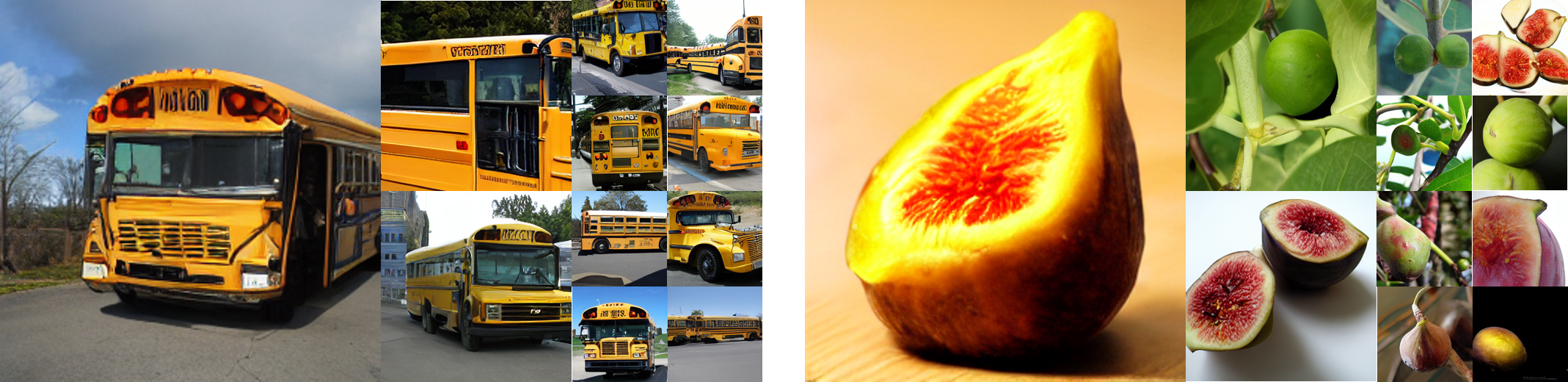}}\\ 
\end{center}
\caption{\textbf{More random samples in specific classes from DPT.}}
\label{fig:moresamples}
\vspace{-.3cm}
\end{figure*}

Fig.~\ref{fig:moresamples} shows more random samples generated by DPT, which are natural, diverse, and semantically consistent with the corresponding classes.

\begin{figure}[t!]
\begin{center}
\subfloat[\emph{One} label per class, \emph{P} (0.93), \emph{R} (0.98)]{\includegraphics[width=.32\columnwidth]{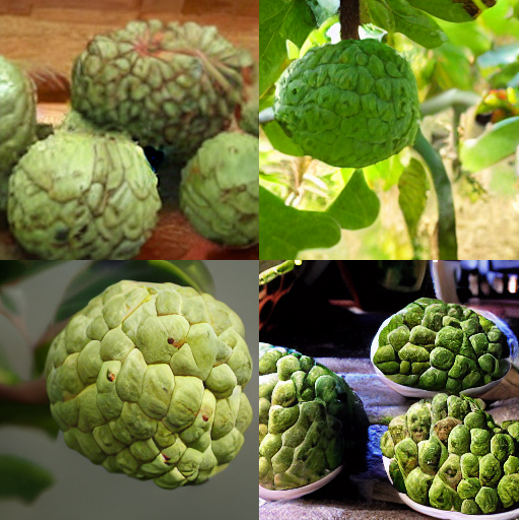}} 
\hspace{0.1cm}\subfloat[ \emph{Two} labels per class, \emph{P} (0.97), \emph{R} (0.97)]
{\includegraphics[width=.32\columnwidth]{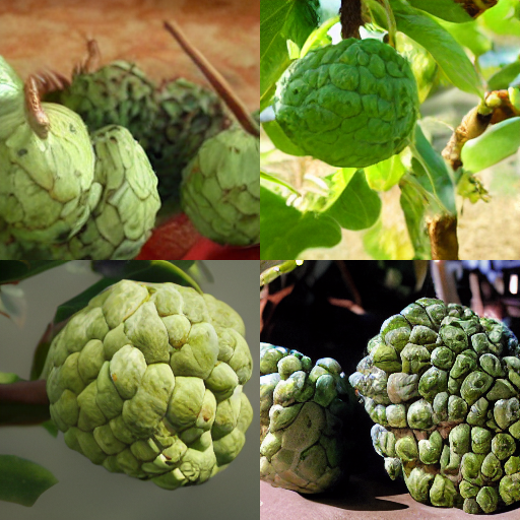}} 
\hspace{0.1cm}\subfloat[ \emph{Five} labels per class, \emph{P} (0.97), \emph{R} (0.97)]
{\includegraphics[width=.32\columnwidth]{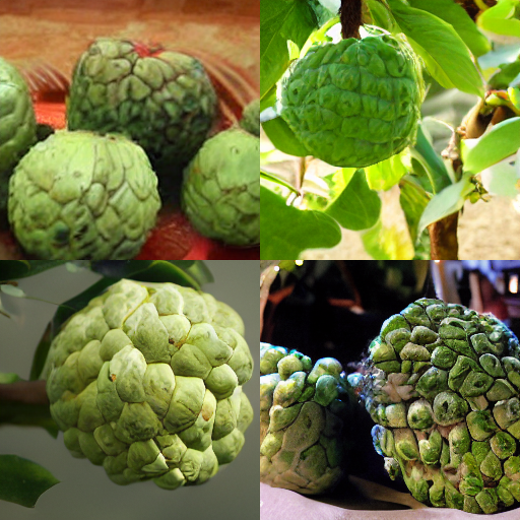}} \\
\subfloat[\emph{One} label per class, \emph{P} (0.08), \emph{R} (0.00)]{\includegraphics[width=.32\columnwidth]{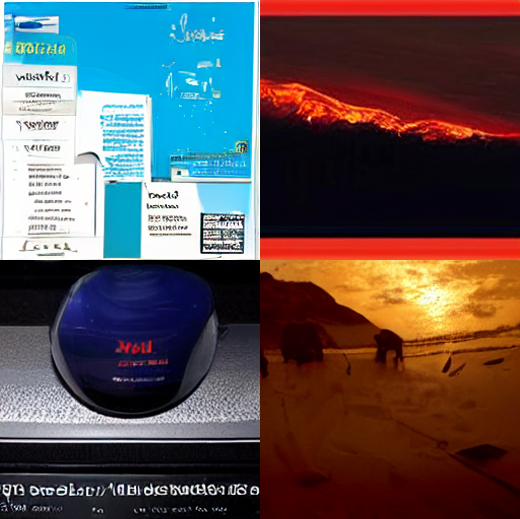}} 
\hspace{0.1cm}\subfloat[ \emph{Two} labels per class, \emph{P} (0.79), \emph{R} (0.85)]
{\includegraphics[width=.32\columnwidth]{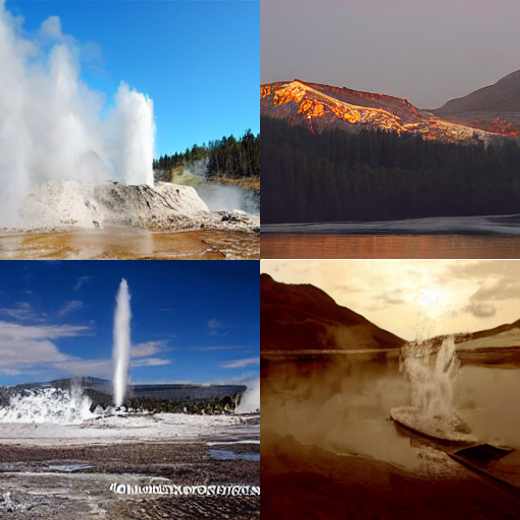}} 
\hspace{0.1cm}\subfloat[ \emph{Five} labels per class, \emph{P} (0.94), \emph{R} (0.99)]
{\includegraphics[width=.32\columnwidth]{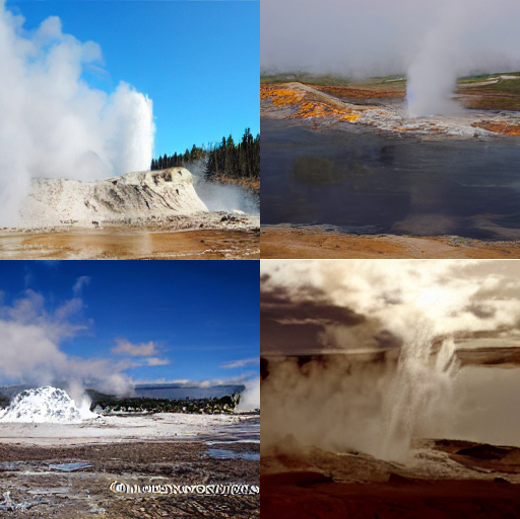}} \\
\subfloat[\emph{One} label per class, \emph{P} (0.02), \emph{R} (0.02)]{\includegraphics[width=.32\columnwidth]{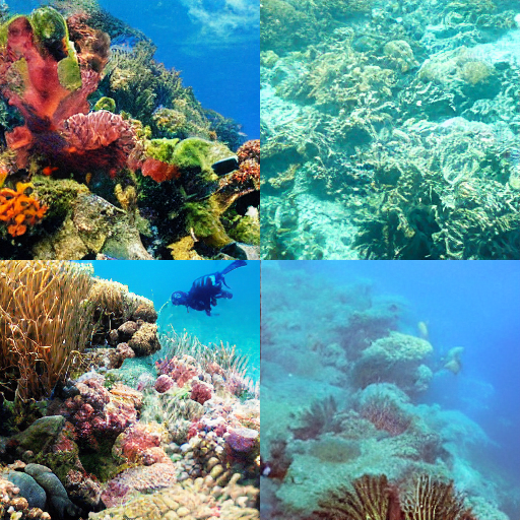}} 
\hspace{0.1cm}\subfloat[ \emph{Two} labels per class, \emph{P} (0.60), \emph{R} (0.98)]
{\includegraphics[width=.32\columnwidth]{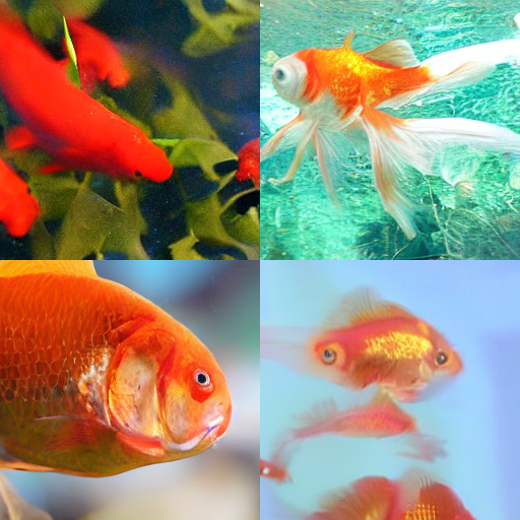}} 
\hspace{0.1cm}\subfloat[ \emph{Five} labels per class, \emph{P} (0.65), \emph{R} (0.98)]
{\includegraphics[width=.32\columnwidth]{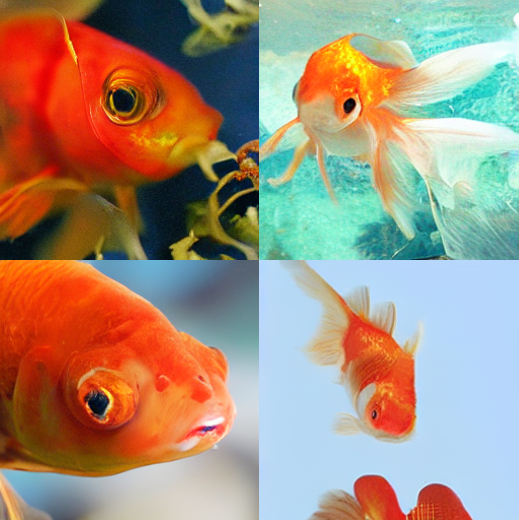}}

\end{center}
\vspace{-.3cm}
\caption{\textbf{
Random samples by varying the number of real labels in the first stage.} More real labels result in smaller noise in pseudo-labels and samples of better visual quality and correct semantics. \emph{Top:} ``Custard apple''. \emph{Middle:} ``Geyser''. \emph{Bottom:} ``Goldfish''.}
\label{fig:accuracy_change_for_generation}
\end{figure}

Moreover, Fig.~\ref{fig:accuracy_change_for_generation} depicts the randomly generated images in selected classes, from DPT trained with one, two, and five real labels per class. As shown in Fig.~\ref{fig:accuracy_change_for_generation} (a), if the classifier can make accurate predictions given one label per class, then DPT can generate images of high quality. However, we find failure cases of DPT in Fig.~\ref{fig:accuracy_change_for_generation} (d) and (g), where the samples are of incorrect semantics due to the large noise in the pseudo-labels. Nevertheless, as the number of labels increases, the generation performance of DPT becomes better due to more accurate predictions.

Notably, in Fig.~\ref{fig:accuracy_change_for_generation}, we fix the same random seed for image generation in the same class across DPT with a different number of labels (e.g., Fig.~\ref{fig:accuracy_change_for_generation} (a-c)) for a fair and clear comparison. The samples of different models given the same random seed are similar because all models attempt to characterize the same diffusion ODE and the discretization of the ODE does not introduce extra noise~\citep{lu2022dpm}, as observed in existing diffusion models~\citep{song2020score,bao2022all}.

\newpage
\subsection{How to use pseudo images in Semi-ViT}
\label{app:different_settings_semi_vit}
\begin{figure}
\centering
\subfloat[Top-1 Accuracy]{\includegraphics[width=.45\columnwidth]{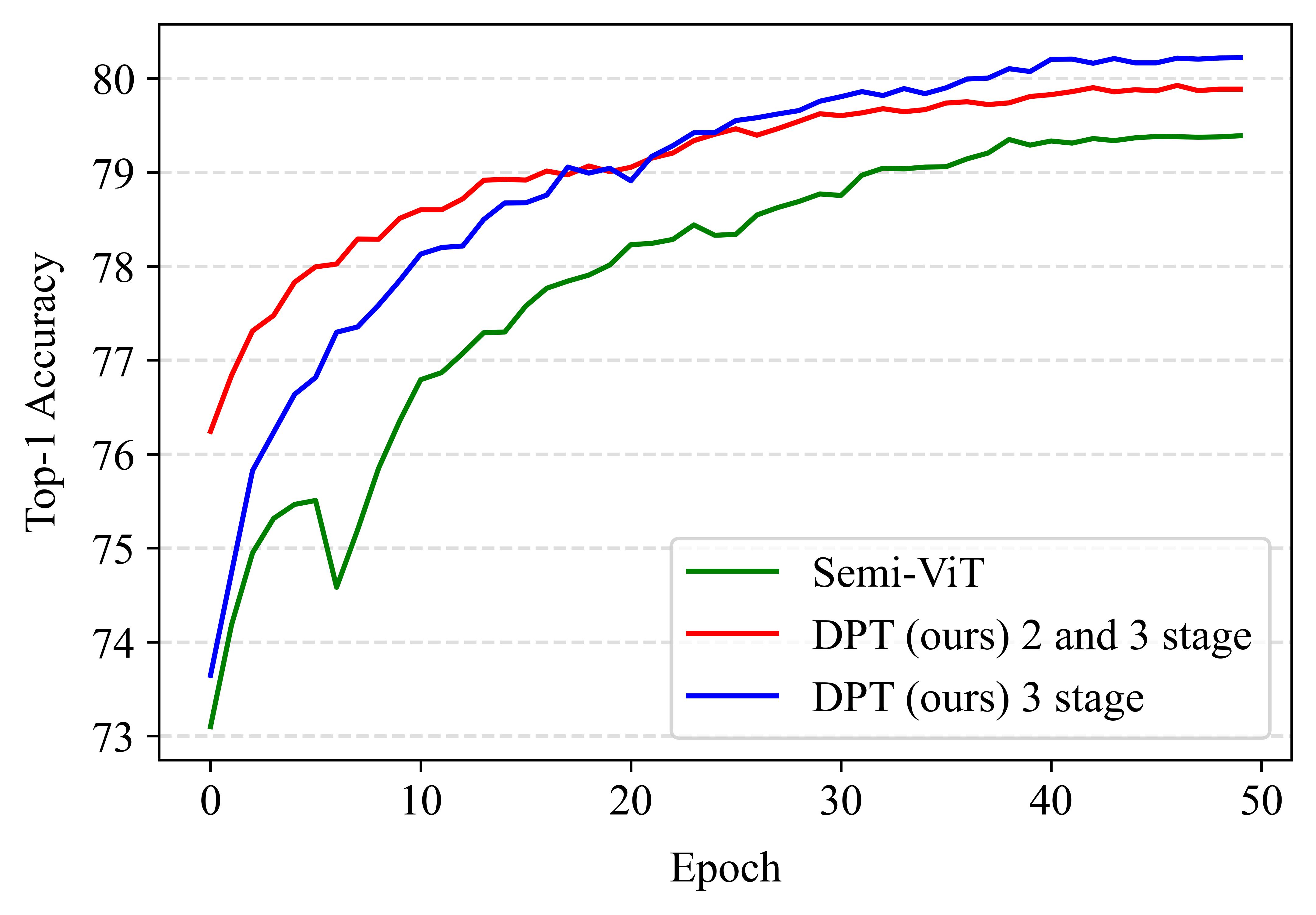}}
\hspace{0.01cm}
\subfloat[Top-5 Accuracy]{\includegraphics[width=.45\columnwidth]{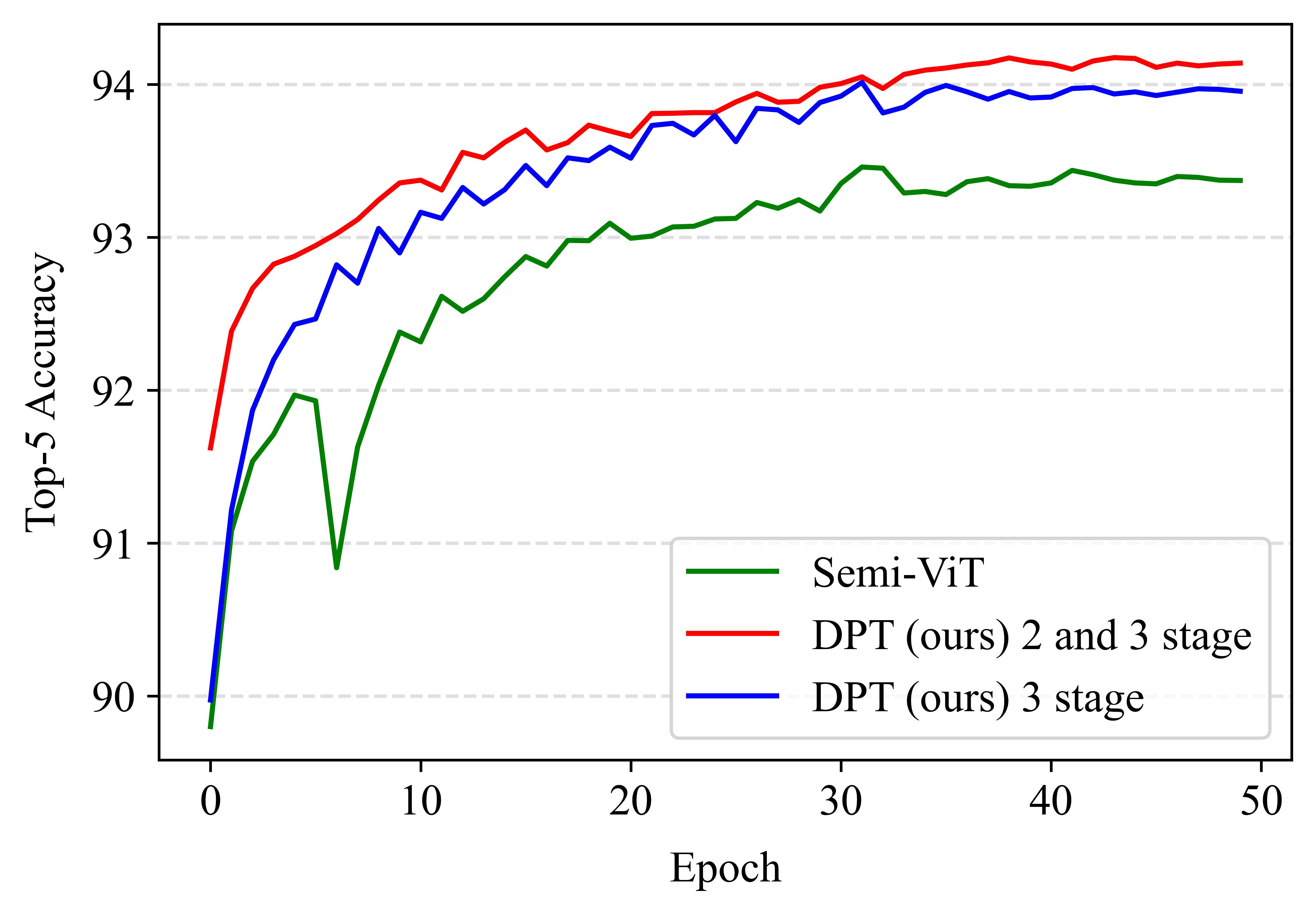}}

\vspace{-.2cm}
\caption{\textbf{Semi-ViT based DPT.} 2 and 3 stage means that we replaces $\mathcal{S}$ with $\mathcal{S} \cup \mathcal{S}_2$ in the two and third stage of Semi-ViT. 3 stage means that we replaces $\mathcal{S}$ with  $\mathcal{S} \cup \mathcal{S}_2$ in the third stage of Semi-ViT. \textbf{(a-b)} These two settings both improve the performance of Semi-ViT and stabilize the training.}
\label{fig:semi_vit_result}
\end{figure}

For Semi-ViT based DPT, in order to fully leverage the pseudo images, we consider the two settings: (1) replaces $\mathcal{S}$ with  $\mathcal{S} \cup \mathcal{S}_2$ in the third stage of Semi-ViT, which is mainly considered in the main text. (2) replaces $\mathcal{S}$ with  $\mathcal{S} \cup \mathcal{S}_2$ in the two and third stages of Semi-ViT. As shown in Fig.~\ref{fig:semi_vit_result}, pseudo images indeed improve the performance of Semi-ViT. Besides, we also find that although the utilization of pseudo images in the second stage of Semi-ViT can provide initial points with high classification accuracy for the third stage, the final top-1 accuracy is lower than just utilizing pseudo images in the third stage of Semi-ViT. 

We also compare DPT with Semi-ViT and state-of-the-art fully supervised models (see Tab.~\ref{tab:compare_supervised}) and find that DPT performs comparably to Inception-v4 \citep{szegedy2017inception}, using only 1$\%$ labels.

\begin{table*}
\caption{\textbf{Comparison with the state-of-the-art fully supervised models on ImageNet classification}  $^\dagger$ labels the results taken from corresponding references and $^\star$ labels the baselines reproduced by us.}
\vskip 0.15in
\small
\centering
\begin{tabular}{llcc}
\toprule
{Method} & Data & Top-1 & Top-5 \\
\midrule
ResNet-50~\citep{he2016deep}$^\dagger$ & ImageNet & 76.0 & 93.0 \\
ResNet-152~\citep{he2016deep}$^\dagger$ & ImageNet & 77.8 & 93.8 \\
Inception-v3~\citep{szegedy2016rethinking}$^\dagger$ & ImageNet & 78.8 & 94.4 \\
Inception-v4~\citep{szegedy2017inception}$^\dagger$ & ImageNet & 80.0 & 95.0 \\
SENet-154~\citep{hu2018squeeze}$^\dagger$ & ImageNet & 81.3 & 95.5 \\
EfficientNet-L2~\citep{tan2019efficientnet}$^\dagger$ & ImageNet & 85.5 & 97.5 \\
\midrule
DeiT-B~\citep{touvron2021training}$^\dagger$ & ImageNet & 81.8 & - \\
Swin-B~\citep{liu2021swin}$^\dagger$ & ImageNet & 83.3 & - \\
MAE~\citep{DBLP:conf/cvpr/HeCXLDG22MAE}$^\dagger$ & ImageNet & 86.9 & - \\
\midrule
Semi-ViT~\citep{cai2022semi}$^\dagger$ & 1$\%$~ImageNet & 80.0 & 93.1 \\
Semi-ViT~\citep{cai2022semi}$^\star$ & 1$\%$~ImageNet & 79.4 & 93.4 \\
DPT~(ours) & 1$\%$~ImageNet & 80.2 & 94.0 \\
\bottomrule
\end{tabular}
\vspace{-.4cm}
\label{tab:compare_supervised} 
\end{table*}

\subsection{Results with More Stages}
\label{app:more_stages}

According to Tab.~\ref{tab:sota_more_semi}, we find that using generative augmentation can lead to more accurate predictions on unlabeled images. Therefore, we attempt to add a further stage employing the refined classifier to predict pseudo-labels for all data, and then re-train the conditional generative model on them. As shown in Tab.~\ref{tab:refined_pseudo_label_in_generation},  these refined pseudo-labels indeed bring a consistent improvement on all quantitative metrics, showing promising promotion of more-stage training. However, note that re-training the conditional generative model is time-consuming and we focus on the three-stage
strategy in this paper for simplicity and efficiency.

\begin{table}[!htb]
\caption{\label{tab:refined_pseudo_label_in_generation} \textbf{Effect of refined pseudo-labels.} Results on ImageNet $256\times256$ benchmark with one label per class. }
\vskip 0.15in
\small
\centering
\begin{tabular}{lccccc}
\toprule
Method  & FID$\downarrow$ & sFID $\downarrow$ & IS $\uparrow$ & Precision $\uparrow$ & Recall $\uparrow$\\
\midrule
DPT  &4.34  & 6.68 & 162.96 & 0.80 & 0.53 \\
DPT with refined pseudo-labels &4.00   & 6.56   & 178.05   & 0.81   & 0.53 \\
\bottomrule
\end{tabular}
\end{table}

\subsection{Can DPT Improve the Upper Bound of Generation Quality}
When all labels are available, the second stage of DPT becomes equivalent to training a supervised diffusion model with real labels. This is essentially the same as a supervised conditional baseline. Therefore, by combining fully supervised labeled data, DPT will not surpass the baseline (e.g. 2.29 FID of U-ViT).

\section{Ablation Studies}
\label{app:ablation_study}
\begin{figure}[!htb]
\centering
\subfloat[]{\includegraphics[width=.32\columnwidth]{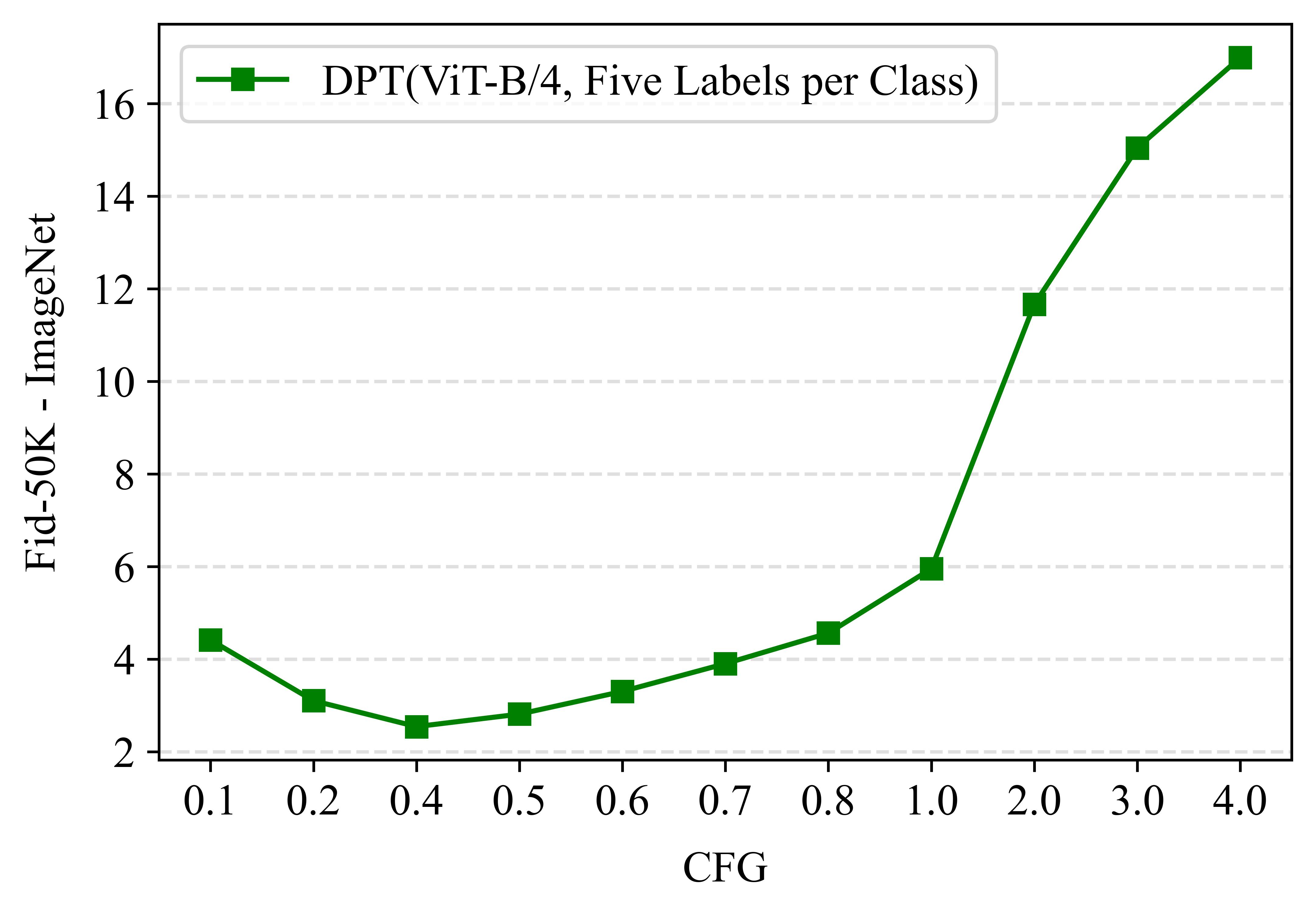}}
\hspace{0.01cm}
\subfloat[]{\includegraphics[width=.32\columnwidth]{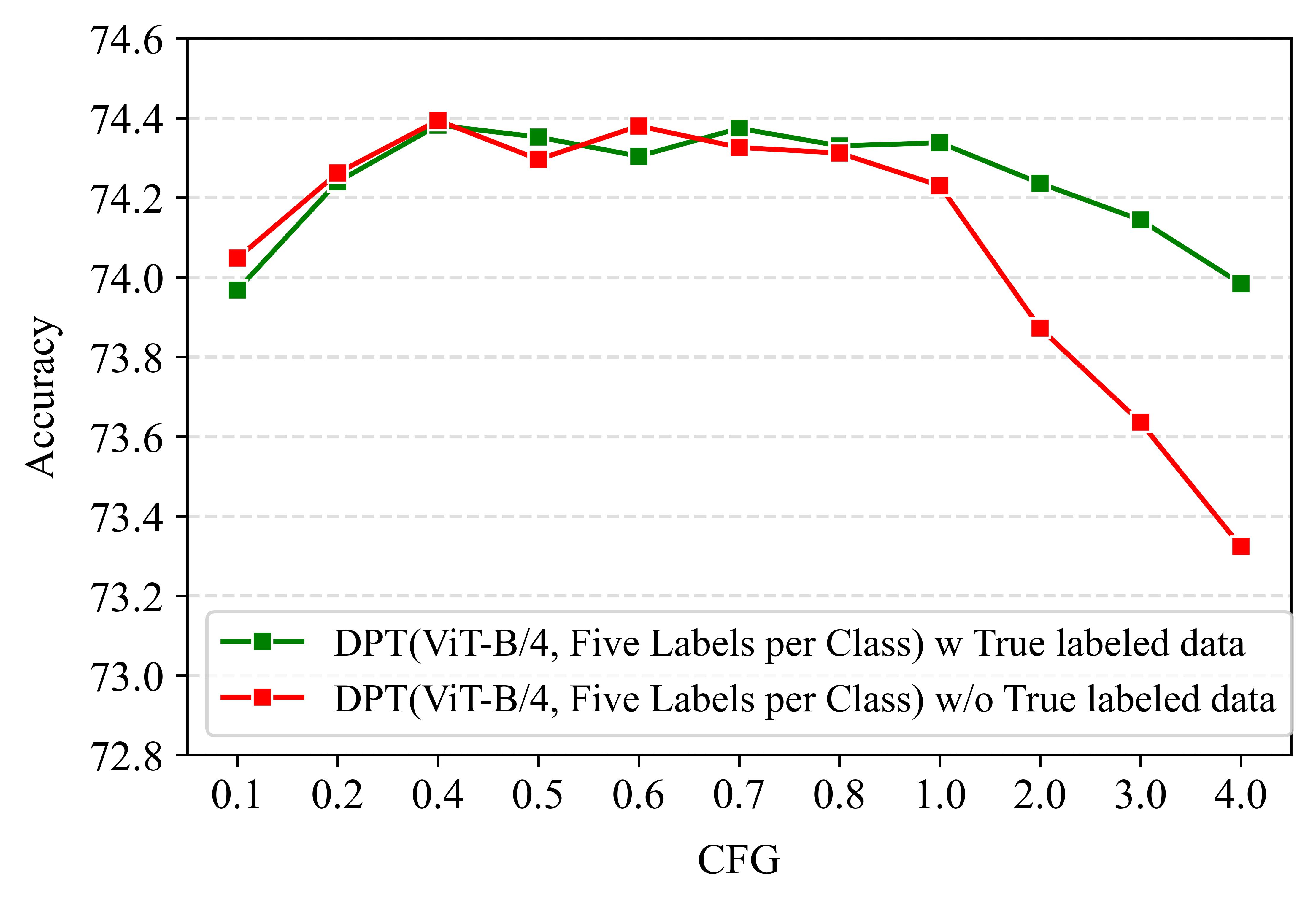}}
\hspace{0.01cm}
\subfloat[]{\includegraphics[width=.32\columnwidth]{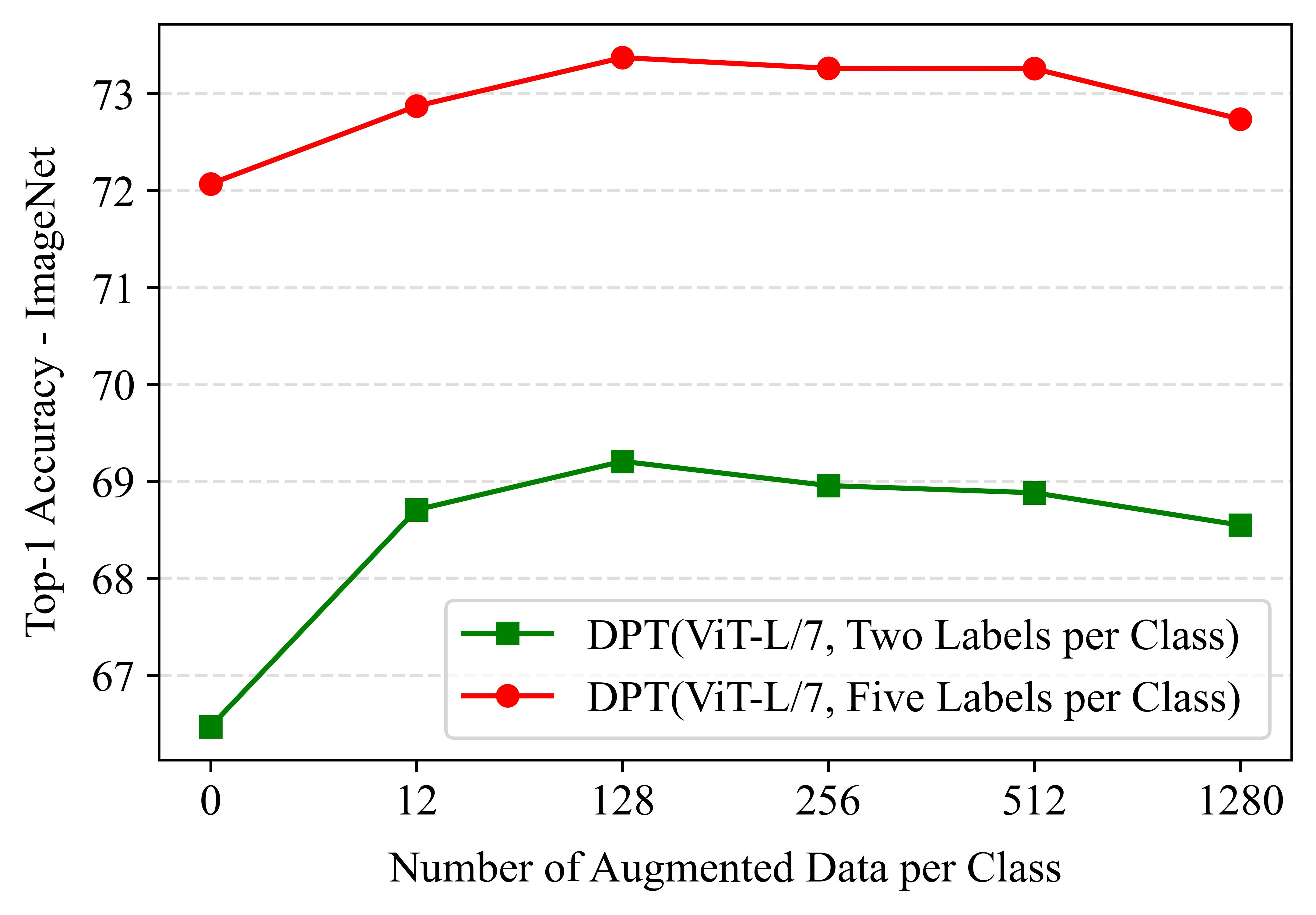}}

\vspace{-.2cm}
\caption{\textbf{Sensitivity.} \textbf{ (a-b)} $CFG$ is highly non-trivial for FID-50K and accuracy. When choosing the $CFG$ that minimizes FID, accuracy tends to be higher, and to some extent, the higher the FID, the worse the accuracy will be. For ImageNet $256\times256$, $CFG=0.4$ is the best choice. \textbf{ (c)} DPT improves the baselines ($K=0$) with $K$ in a large range. $K=128$ is the best choice.}
\label{fig:sensitivity}
\end{figure}

\textbf{Sensitivity of \emph{K}.} The most important hyperparameter in DPT is the number of augmented pseudo images per class, i.e., $K$. In order to analyze the sensitivity of DPT with respect to $K$, we perform a simple grid search on $\{12, 128, 256, 512, 1280\}$ in MSN (ViT-L/7) with two and five labels per class and find that $K=128$ is the best choice. Therefore, we set $K=128$ as the default value if not specified (see Fig.~\ref{fig:sensitivity} (c)). We observed that $K=128$ was the optimal choice in both settings. Intuitively, an overly large $K$ would cause the classifier to be dominated by pseudo images and ignore real data, which explains the sub-optimal performance with $K>128$. Nevertheless, according to Fig.~\ref{fig:sensitivity} (c), DPT consistently and substantially improved the baselines ($K=0$) over a large range of values in \{12, 128, 256, 512, 1280\}.

\textbf{Sensitivity of \emph{CFG}.} In the third stage of DPT, we replaces $\mathcal{S}$ with $\mathcal{S} \cup \mathcal{S}_2$. The choice of \emph{w} is highly non-trivial for the quality of pseudo images, Therefore, it is also significant for classification. We conducted experiments on the ImageNet dataset with five labels per class, sweeping over a range of values for \emph{w} from 0.1 to 4.0, and evaluated the performance of the model in terms of FID-50K and top-1 Accuracy. The results are presented in Figure~\ref{fig:sensitivity} (a-b). Moreover, we find that the choice of CFG that minimizes FID will lead to the best accuracy. Specifically, we find that $CFG=0.4$ achieves the best performance for ImageNet $256\times256$, while $CFG=0.8$ and $CFG=0.7$ are the optimal choices for ImageNet $128\times128$ and $512\times512$, respectively.

\section{How Does Classifier Benefit Generation?}
\label{app:class_wise_analysis_classification_help_generation}
\begin{figure}[t!]
\begin{center}
\subfloat[\emph{P}(0.99),\emph{R}(1.00) ``Giant Panda''
]{\includegraphics[width=.32\columnwidth]{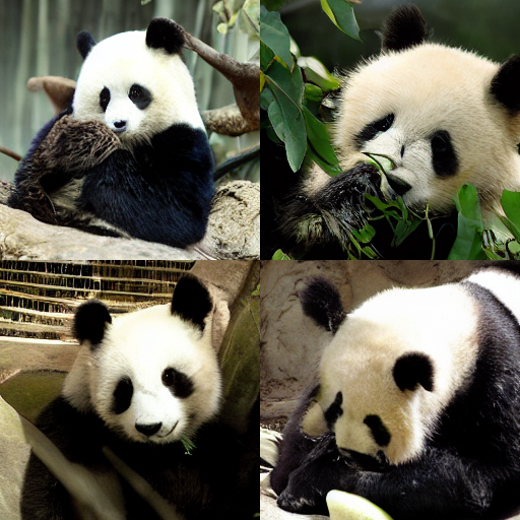}} 
\hspace{0.1cm}\subfloat[\emph{P}(0.16),\emph{R}(0.75) ``Purse''
]{\includegraphics[width=.32\columnwidth]{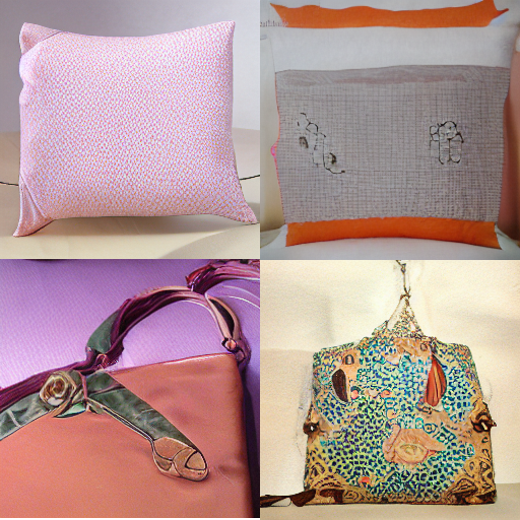}} 
\hspace{0.1cm}\subfloat[\emph{P}(0.99),\emph{R}(0.23) ``Hornbill''
]{\includegraphics[width=.32\columnwidth]{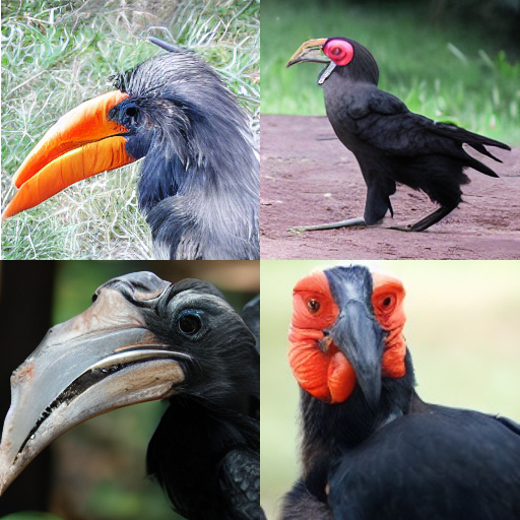}} 
\end{center}
\vspace{-.3cm}
\caption{\textbf{Random samples in selected classes with different \emph{P} and \emph{R}.} (a) High \emph{P} and \emph{R} ensure  good samples. (b) Low \emph{P} leads to semantical confusion. (c) Low \emph{R} lowers the visual quality.}
\label{fig:precision_recall_analysis_for_generation}
\end{figure}

\begin{figure}[t!]
\begin{center}
\subfloat[Distribution of \emph{R}]
{\includegraphics[width=0.34\columnwidth]{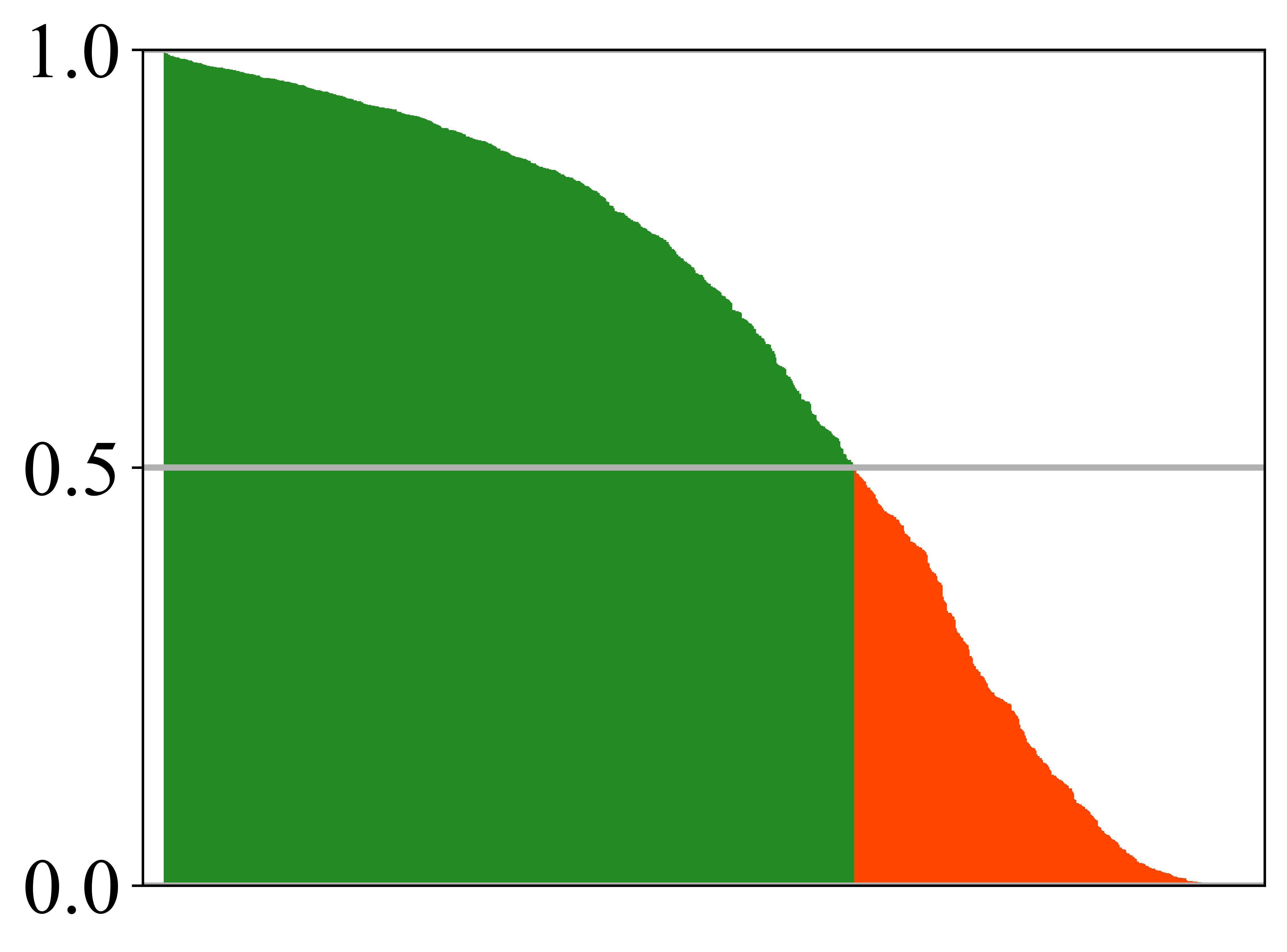}}
\vspace{.1cm}\subfloat[Distribution of  \emph{P}]{\includegraphics[width=0.34\columnwidth]{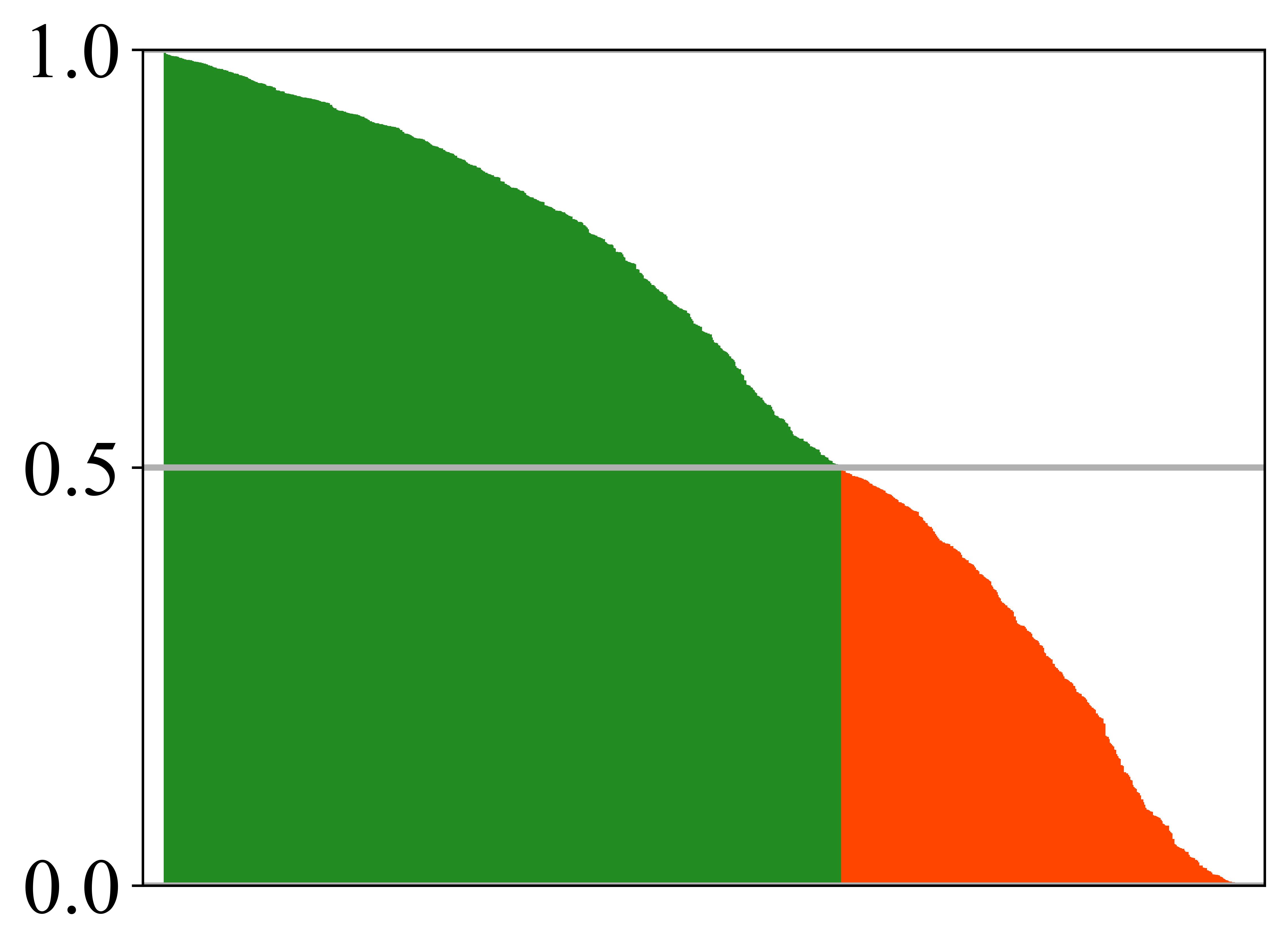}}
\end{center}
\vspace{-.5cm}
\caption{\textbf{Distributions of \emph{R} and \emph{P}}. The vertical axis represents the values of \emph{P} and \emph{R} w.r.t the classifier trained in the first stage. The horizontal axis represents all classes sorted by the values.}
\label{fig:r_p_distribution}
\end{figure}

We explain why the classifier can benefit the generative model through class-level visualization and analysis based on precision and recall on training data.
For a given class $y$, the precision and recall w.r.t. the classifier is defined by \emph{P = TP/(TP + FP)}\footnote{We omit the dependence on $y$ for simplicity.} and \emph{R = TP/(TP + FN)}, where \emph{TP}, \emph{FP}, and \emph{FN} denote the number of true positive, false positive, and false negative samples respectively. Intuitively, higher \emph{P} and \emph{R} suggest smaller noise in pseudo-labels and result in better samples. Therefore, we employ strong semi-supervised learners~\citep{assran2022masked} in the first stage to reduce the noise. 

We select three representative classes with different values of \emph{P} and \emph{R} and visualize the samples in Fig.~\ref{fig:precision_recall_analysis_for_generation}. In particular, the pseudo-labels in a class with both high \emph{P} and \emph{R} contain little noise, leading to good samples (Fig.~\ref{fig:precision_recall_analysis_for_generation} (a)).
In contrast, on one hand, a low \emph{P} means that a large fraction of images labeled as $y$ in $\mathcal{S}_1$ are not actually in class $y$, and the samples from the generative model given the label $y$ can be semantically wrong (Fig.~\ref{fig:precision_recall_analysis_for_generation} (b)). On the other hand, a low \emph{R} means that a large fraction of images in class $y$ are misclassified as other labels and the samples  from the generative model can be less realistic due to the lack of 
training images in class $y$ (Fig.~\ref{fig:precision_recall_analysis_for_generation} (c)). 

Through the analysis of the three representative classes above, the classifier benefits the generator by bringing more accurate and low-noise pseudo labels to the generator. In particular, with \emph{one label per class}, MSN~\citep{assran2022masked} with ViT-L/7 achieves a top-1 training accuracy of 60.3\%. As presented in Fig.~\ref{fig:r_p_distribution}, \emph{R} and \emph{P} of most classes are higher than 0.5. Quantitatively, despite using only $<0.1\%$ of the labels, DPT achieves an FID of 3.08, compared to the FID of 2.29 achieved by U-ViT-Huge with full labels. The reduction in FID is not significant. This demonstrates that although noise exists, such a strong classifier can benefit the generative model overall and reduce the usage of labels.

\section{How Does Generative Model Benefit Classification?}\label{app:analysis_generation_help_classification}

Similarly to Appendix.~\ref{app:class_wise_analysis_classification_help_generation}, we explain
why the generative model can benefit the classifier through
class-level visualization and analysis based on precision (\emph{P}) and
recall (\emph{R}). 

\begin{figure*}[t!]
\begin{center}
\subfloat[\emph{R} (0.24 $\rightarrow$ 0.87) ``Albatross'' 
]
{\includegraphics[width=.32\columnwidth]{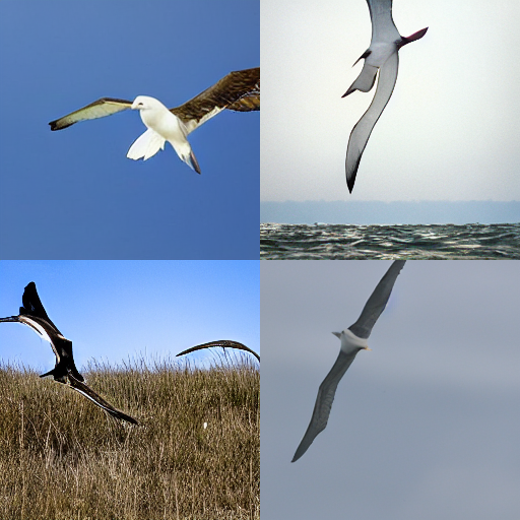}}
\hspace{0.01cm}
\subfloat[\emph{R} (0.22 $\rightarrow$ 0.75) ``Timber wolf''
]
{\includegraphics[width=.32\columnwidth]{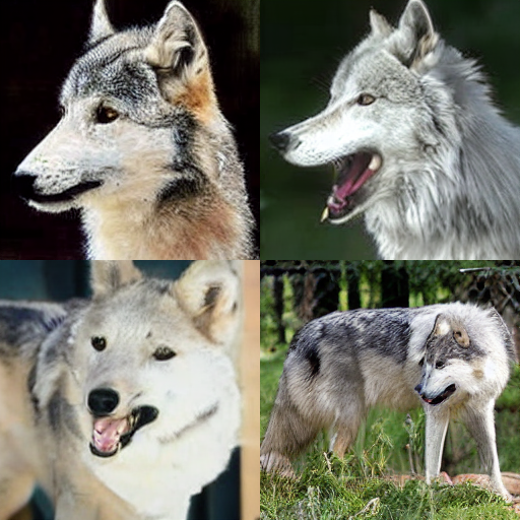}}
\hspace{0.01cm}
\subfloat[\emph{R} (0.26 $\rightarrow$ 0.01) ``Bathtub''
]
{\includegraphics[width=.32\columnwidth]{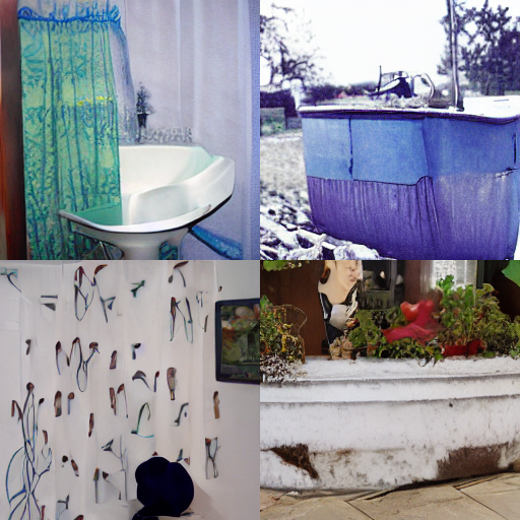}}
\end{center}
\vspace{-.3cm}
\caption{\textbf{Random samples in selected classes with different values of change of \emph{R}.} Values of change are presented in parentheses. (a-b) If pseudo images are realistic and semantically correct, they can  benefit classification. (c) Otherwise, they hurt performance.
}
\vspace{-.2cm}
\label{fig:generation_for_recall_improve}
\end{figure*}

We select three representative classes with different values of change of \emph{R} for visualization in Fig.~\ref{fig:generation_for_recall_improve}. 
If the pseudo images in class $y$ are realistic, diverse, and semantically correct, then it can increase the corresponding \emph{R} as presented in Fig.~\ref{fig:generation_for_recall_improve} (a-b). Instead, poor samples may hurt the classification performance in the corresponding class, as shown in Fig.~\ref{fig:generation_for_recall_improve} (c). 

\begin{figure*}[t!]
\begin{center}
\subfloat[\emph{P} $=0.67$ without pseudo images.]{\includegraphics[height=.30\columnwidth]{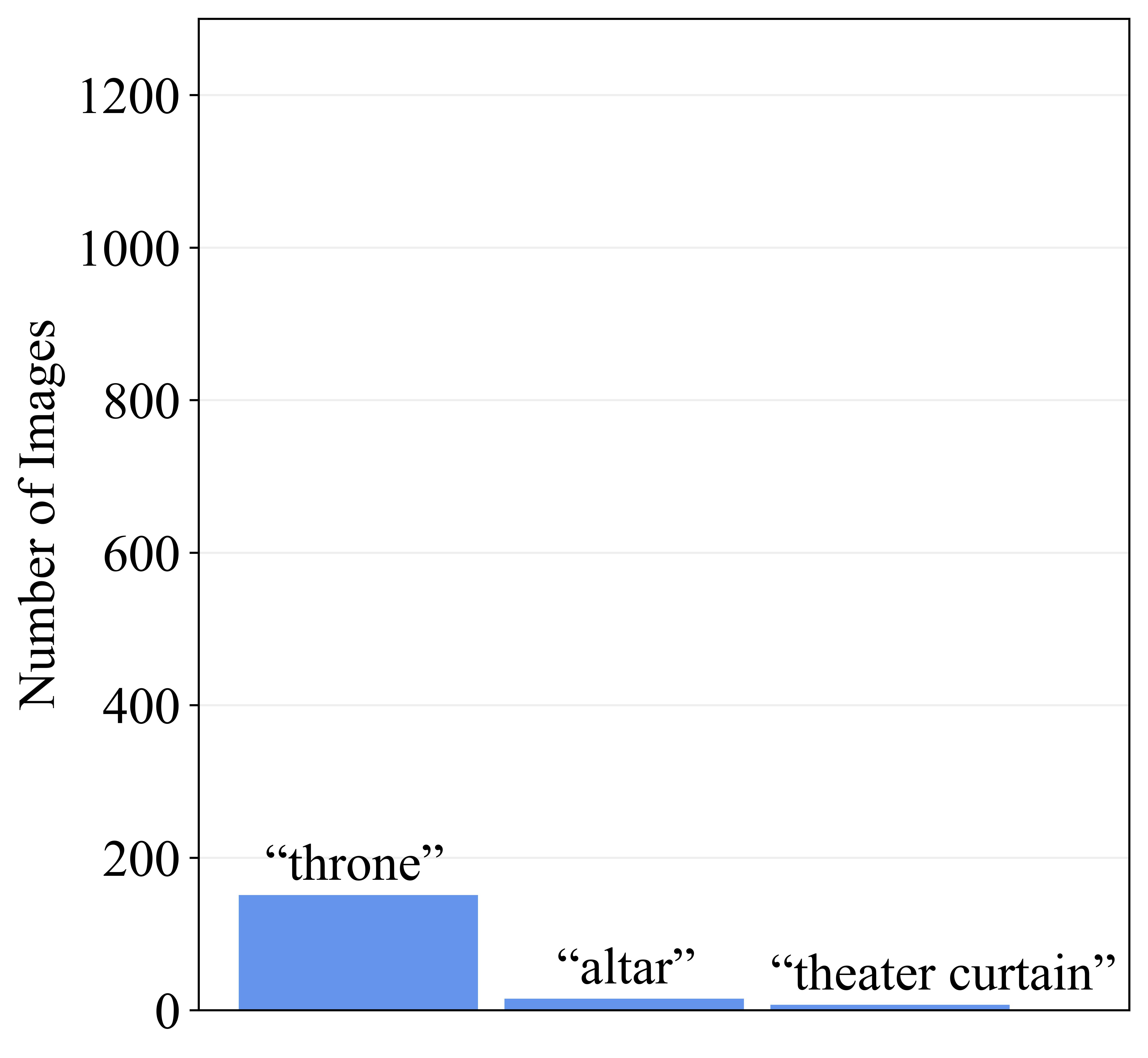}}
\hspace{0.1cm}
\subfloat[\emph{P} $=0.89$ with pseudo images.]{\includegraphics[height=.30\columnwidth]{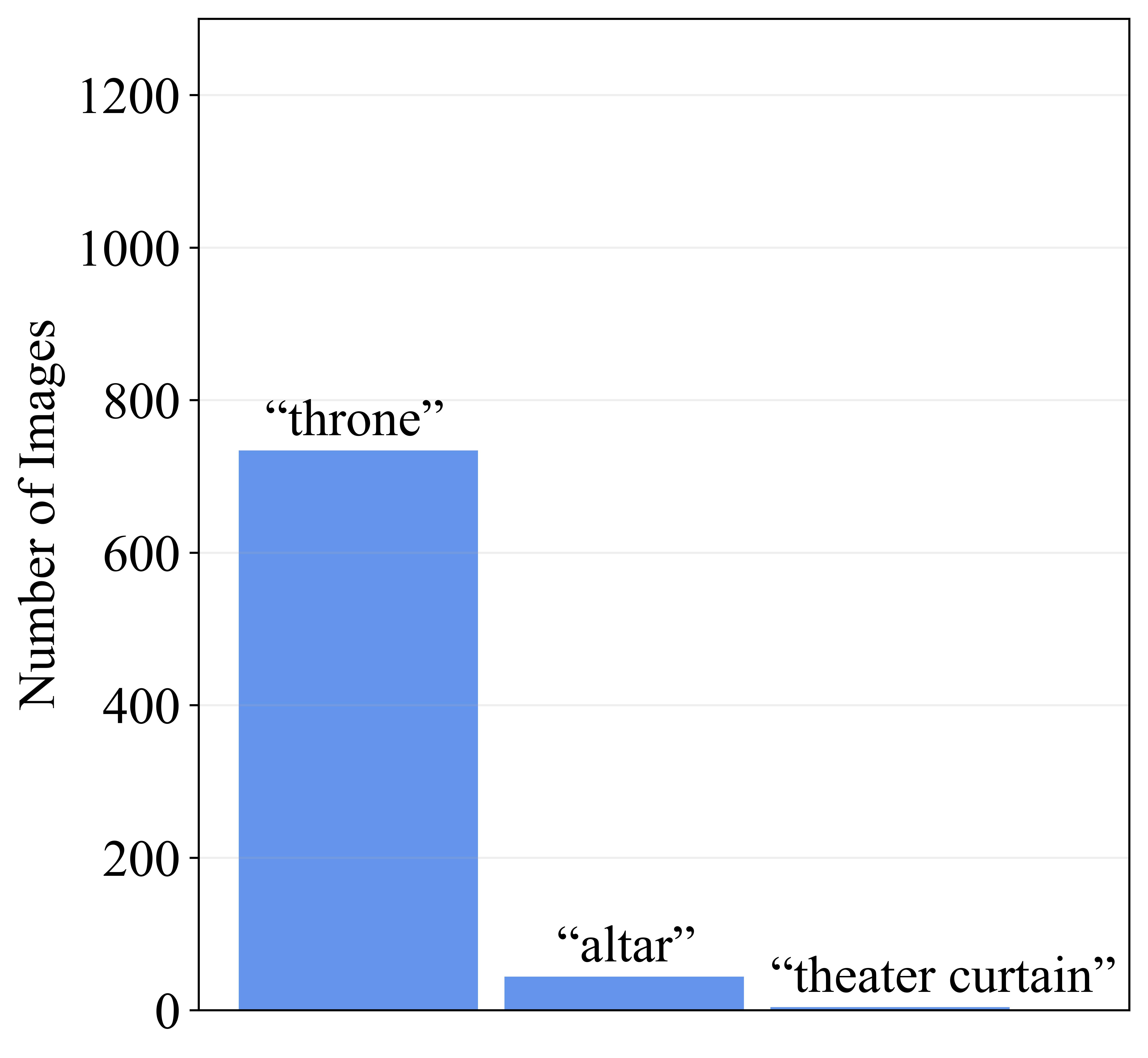}}
\hspace{0.1cm}
\subfloat[``Throne''.]{\includegraphics[height=.30\columnwidth]{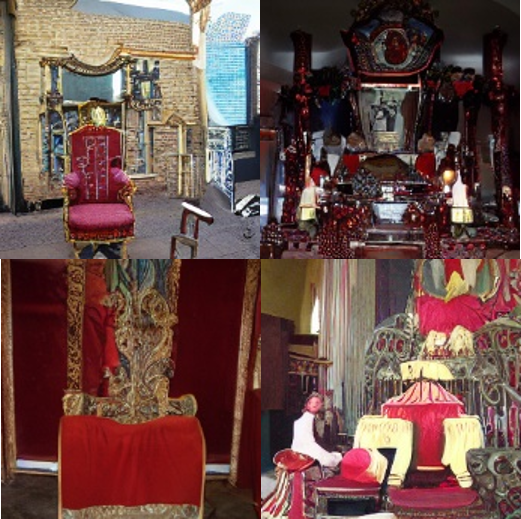}}\\
\subfloat[\emph{P} $=0.52$ without pseudo images.]{\includegraphics[height=.30\columnwidth]{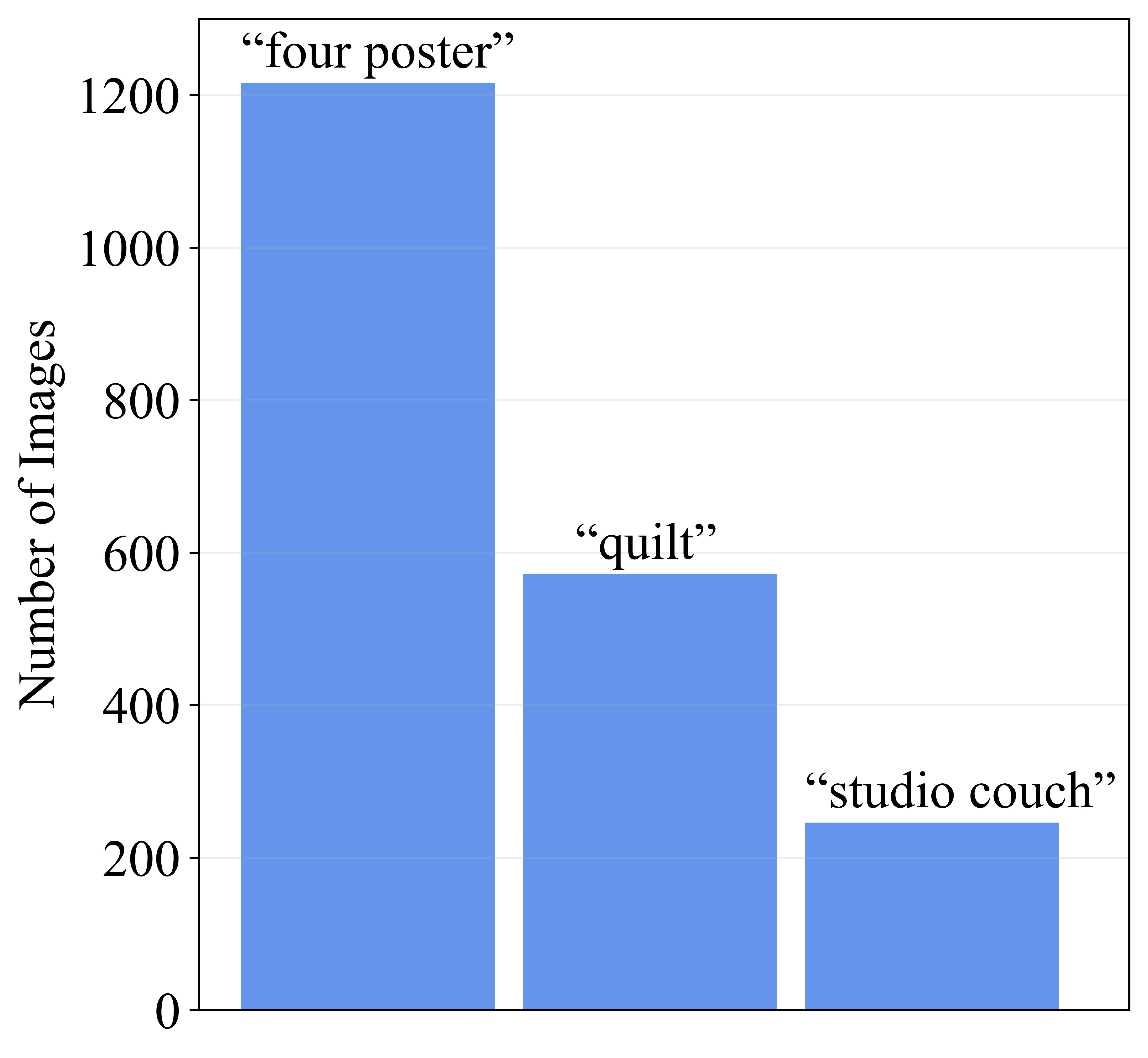}}
\hspace{0.1cm}
\subfloat[\emph{P} $=0.86$ with pseudo images.]{\includegraphics[height=.30\columnwidth]{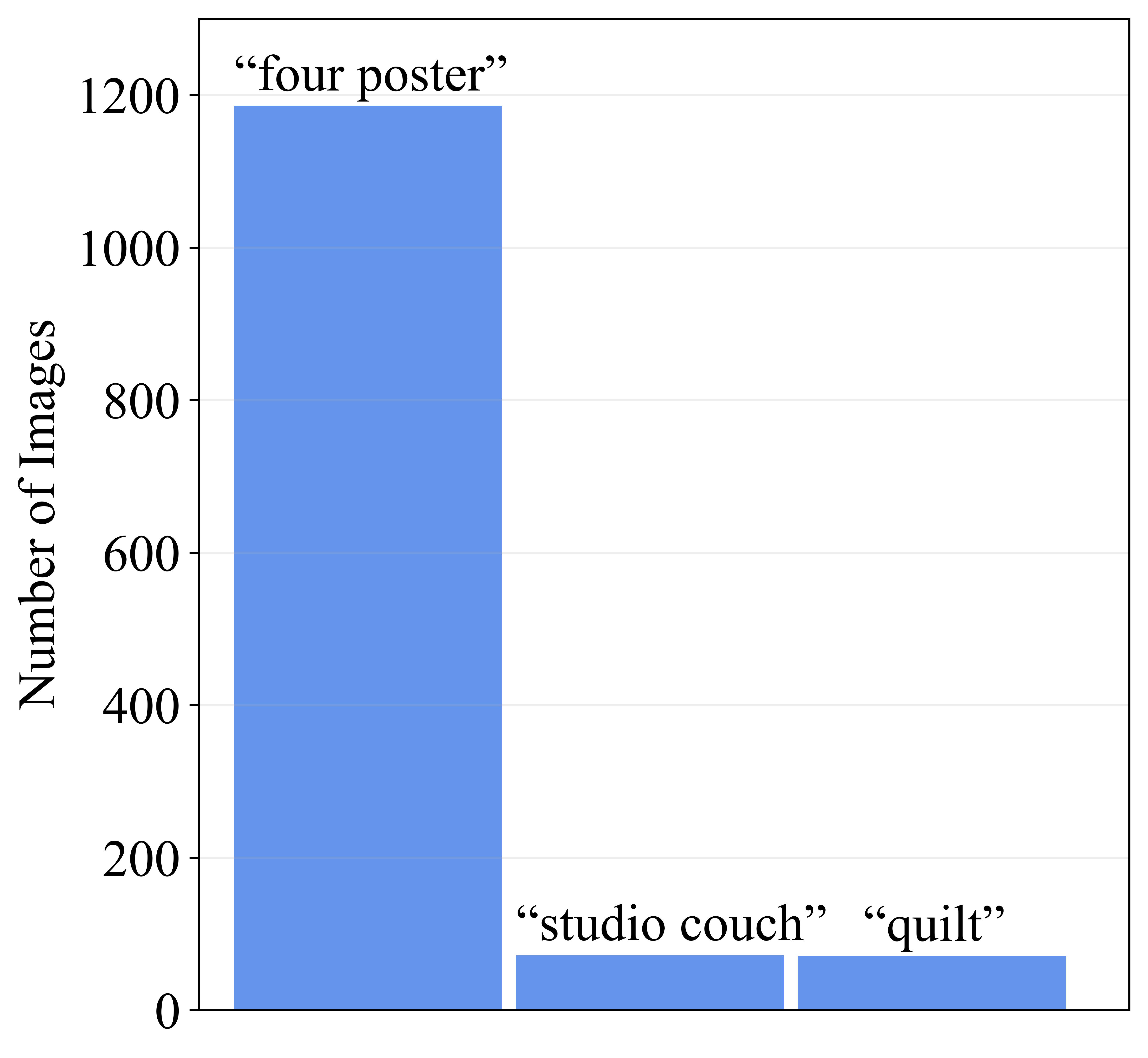}}
\hspace{0.1cm}
\subfloat[\emph{Left:} ``Four poster''. \emph{Right:} ``Quilt''.]{\includegraphics[height=.30\columnwidth]{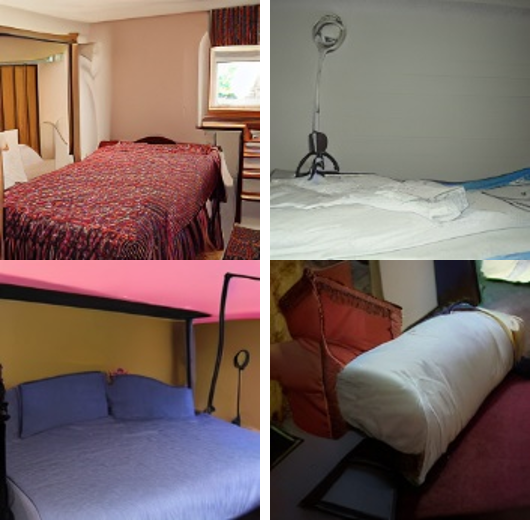}}\\
\end{center}
\caption{\textbf{Detailed analysis in selected classes with a positive change of \emph{P}.}  \emph{Top:} ``Throne''.
High-quality samples in the class ``throne'' (c) directly increase \emph{TP} and improve \emph{P} (a-b). 
\emph{Bottom:} ``Four poster''. The samples in both ``four poster'' and ``quilt'' are of high quality (f). The classifier reduces \emph{FP} with such pseudo samples and improves \emph{P} (d-e). }
\label{fig:precision_analysis}
\vspace{-.3cm}
\end{figure*}

The analysis of \emph{P} involves pseudo images in multiple classes. According to the definition of precision (i.e. \emph{P = TP/(TP + FP)}), the pseudo images can affect \emph{P} through not only \emph{TP} but also \emph{FP}. We select two representative classes with positive changes of \emph{P} to visualize both cases, as shown in Fig.~\ref{fig:precision_analysis}. We select the top-three classes according to the number of images classified as ``throne'' and present the numbers w.r.t. the classifier after the first and third stages in  Fig.~\ref{fig:precision_analysis} (a) and (b) respectively. As shown in Fig.~\ref{fig:precision_analysis} (c), high-quality samples in the class ``throne'' directly increases \emph{TP} (i.e., the number of images in class ``throne'' classified as ``throne'') and improves \emph{P}. We also present the top-three classes related to ``four poster'' in Fig.~\ref{fig:precision_analysis} (d) and (e). It can be seen that \emph{P} in class ``four poster'' increases because of \emph{FP} (of class ``quilt'' especially) decreases. We visualize random samples in both ``four poster'' and ``quilt'' in Fig.~\ref{fig:precision_analysis} (f). These high-quality samples help the classifier to distinguish the two classes, which explains the change of \emph{FP} and \emph{P}. A similar analysis can be conducted for classes with negative change of \emph{P}.

We mention that we analyze the change of \emph{P} and \emph{R} in the third stage on the training set instead of the validation set. This is because the training set is of a much larger size and therefore leads to a much smaller variance in the estimate of $\emph{P}$ and $\emph{R}$. Since most of the data are unlabeled in the training set, this does not introduce a large bias in the estimate of $\emph{P}$ and $\emph{R}$.

\input{more_pictures}
\end{document}

%% file: more_pictures.tex
\clearpage
\begin{figure}[t!]
\begin{center}
\subfloat[
]{\includegraphics[height=1.05\textwidth]{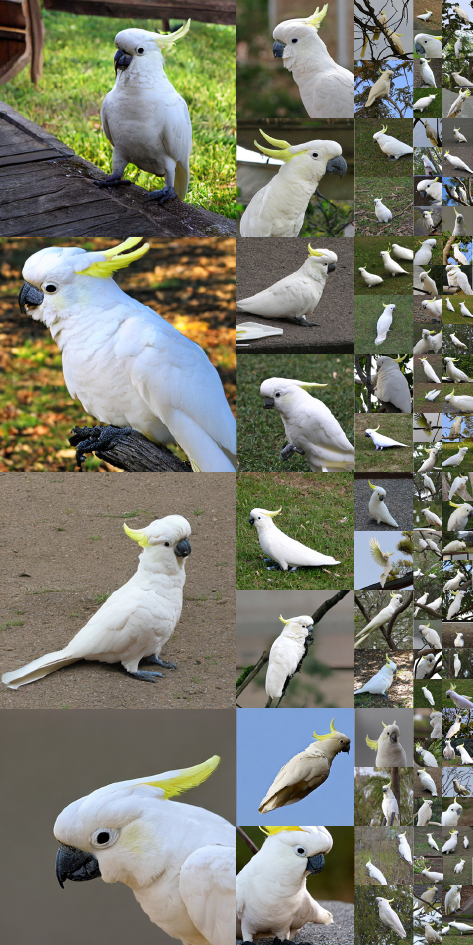}}
\subfloat[]{\includegraphics[height=1.05\textwidth]{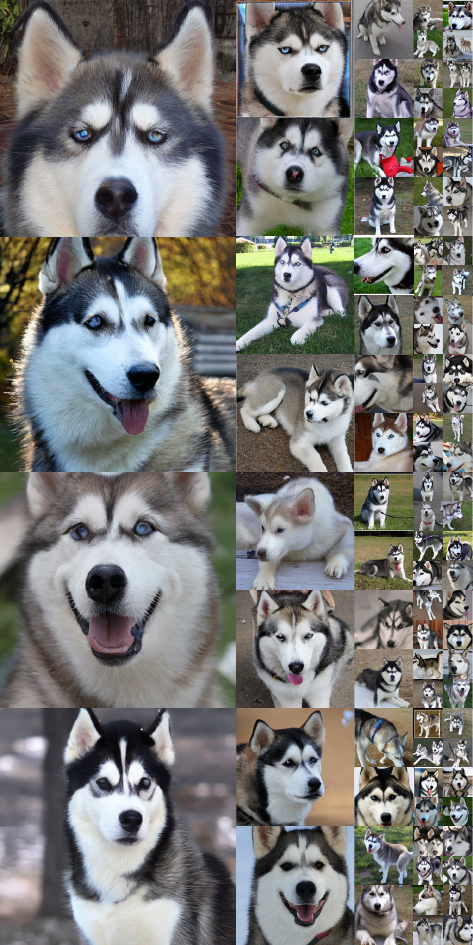}}
\end{center}
\vspace{-.2cm}
\caption{512$\times$512 samples of DPT trained with five labels per class. \emph{CFG} = 3.0}
\vspace{-.3cm}
\end{figure}

\clearpage
\begin{figure}[t!]
\begin{center}
\subfloat[
]{\includegraphics[height=1.05\textwidth]{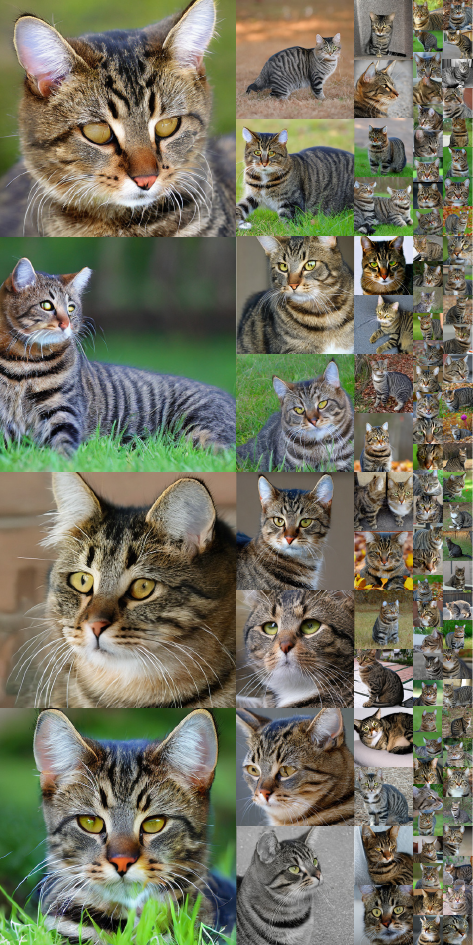}}
\subfloat[]{\includegraphics[height=1.05\textwidth]{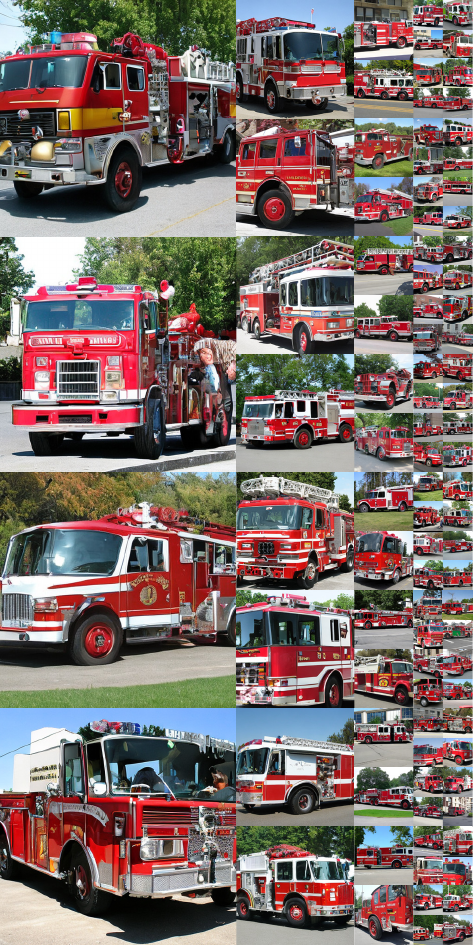}}
\end{center}
\vspace{-.2cm}
\caption{512$\times$512 samples of DPT trained with five labels per class. \emph{CFG} = 3.0}
\vspace{-.3cm}
\end{figure}

\clearpage
\begin{figure}[t!]
\begin{center}
\subfloat[
]{\includegraphics[height=1.05\textwidth]{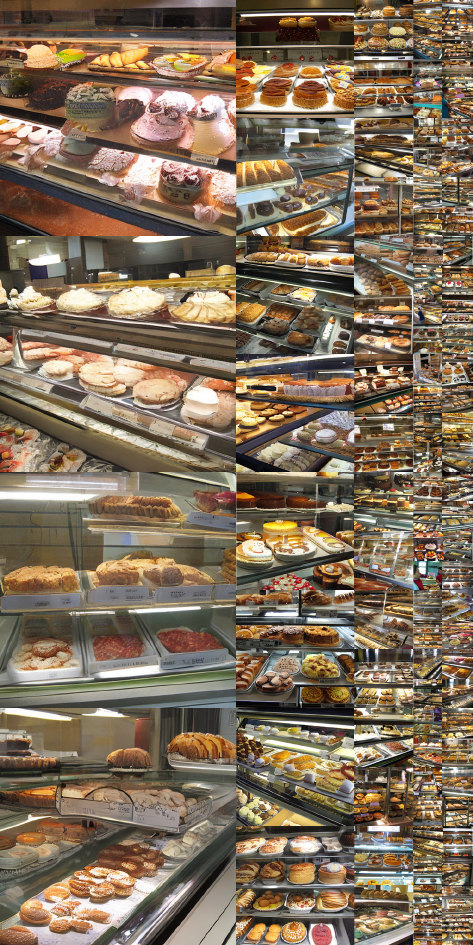}}
\subfloat[]{\includegraphics[height=1.05\textwidth]{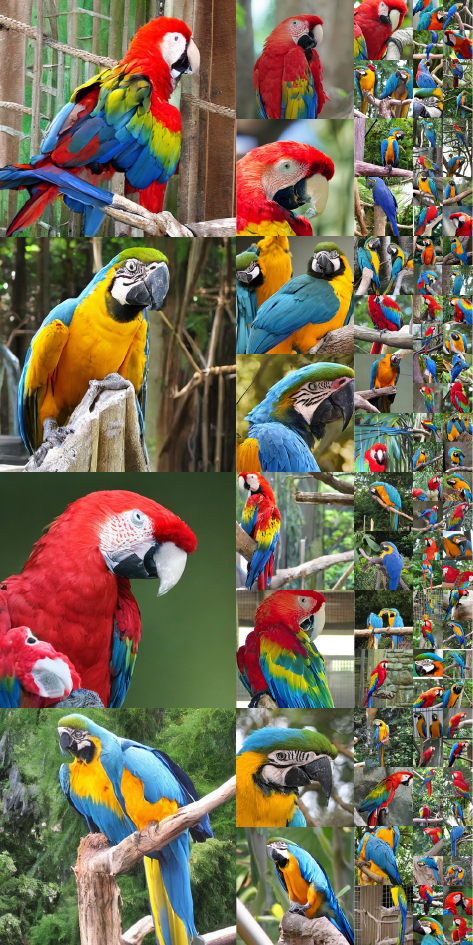}}
\end{center}
\vspace{-.2cm}
\caption{512$\times$512 samples of DPT trained with five labels per class. \emph{CFG} = 3.0}
\vspace{-.3cm}
\end{figure}